\documentclass[3p,authoryear,preprint,12pt]{elsarticle}
\makeatletter\if@twocolumn\PassOptionsToPackage{switch}{lineno}\else\fi\makeatother

\usepackage{tabulary,xcolor}
\usepackage{amsfonts,amsmath,amssymb}
\usepackage[T1]{fontenc}

\usepackage{ifluatex}
\ifluatex
\usepackage{fontspec}
\defaultfontfeatures{Ligatures=TeX}
\usepackage[]{unicode-math}
\unimathsetup{math-style=TeX}
\else 
\usepackage[utf8]{inputenc}
\fi 
\ifluatex\else\usepackage{stmaryrd}\fi

\usepackage{amsmath}
\usepackage{algorithm}
\usepackage{algorithmic}
\usepackage{multirow}
\usepackage{amssymb}
\usepackage{booktabs}
\usepackage{xcolor, soul}
\usepackage[hidelinks]{hyperref}
\numberwithin{equation}{section}

\newcommand{\revise}[1]{\textcolor{black}{#1}}

\usepackage{fancyhdr}
\begin{document}

\begin{frontmatter}



\title{CarcassFormer: An End-to-end Transformer-based Framework for Simultaneous Localization, Segmentation and Classification of Poultry Carcass Defect}


\author[inst1]{Minh Tran$^\dagger$}
\author[inst1]{Sang Truong$^\dagger$ \footnote{ $^\dagger$ indicates the same contribution}}
\author[inst2]{Arthur F. A. Fernandes}
\author[inst3]{Michael T. Kidd}
\author[inst1]{Ngan Le $^\ddagger$ \footnote{ $^\ddagger$ indicates corresponding author}}

\affiliation[inst1]{organization={Department of Computer Science and Computer Engineering},
            addressline={1 University of Arkansas}, 
            city={Fayetteville},
            postcode={72701}, 
            state={Arkansas},
            country={USA}}

\affiliation[inst2]{organization={Cobb Vantress, Inc},
            addressline={4703 US HWY 412 E}, 
            city={Siloam Springs},
            postcode={72761}, 
            state={Arkansas},
            country={USA}}
\affiliation[inst3]{organization={Department of Poultry Science},
            addressline={1260 W. Maple, POSC O-114}, 
            city={Fayetteville},
            postcode={72701}, 
            state={Arkansas},
            country={USA}}

\begin{abstract}
In the food industry, assessing the quality of poultry carcasses during processing is a crucial step. This study proposes an effective approach for automating the assessment of carcass quality without requiring skilled labor or inspector involvement. The proposed system is based on machine learning (ML) and computer vision (CV) techniques, enabling automated defect detection and carcass quality assessment. To this end, an end-to-end framework called CarcassFormer is introduced. It is built upon a Transformer-based architecture designed to effectively extract visual representations while simultaneously detecting, segmenting, and classifying poultry carcass defects. Our proposed framework is capable of analyzing imperfections resulting from production and transport welfare issues, as well as processing plant stunner, scalder, picker, and other equipment malfunctions. 

To benchmark the framework, a dataset of 7,321 images was initially acquired, which contained both single and multiple carcasses per image. In this study, the performance of the CarcassFormer system is compared with other state-of-the-art (SOTA) approaches for both classification, detection, and segmentation tasks. Through extensive quantitative experiments, our framework consistently outperforms existing methods, demonstrating remarkable improvements across various evaluation metrics such as AP, AP@50, and AP@75. Furthermore, the qualitative results highlight the strengths of CarcassFormer in capturing fine details, including feathers, and accurately localizing and segmenting carcasses with high precision. To facilitate further research and collaboration, the source code and trained models will be made publicly available upon acceptance.

\end{abstract}

\begin{graphicalabstract}

\includegraphics[width=\linewidth]{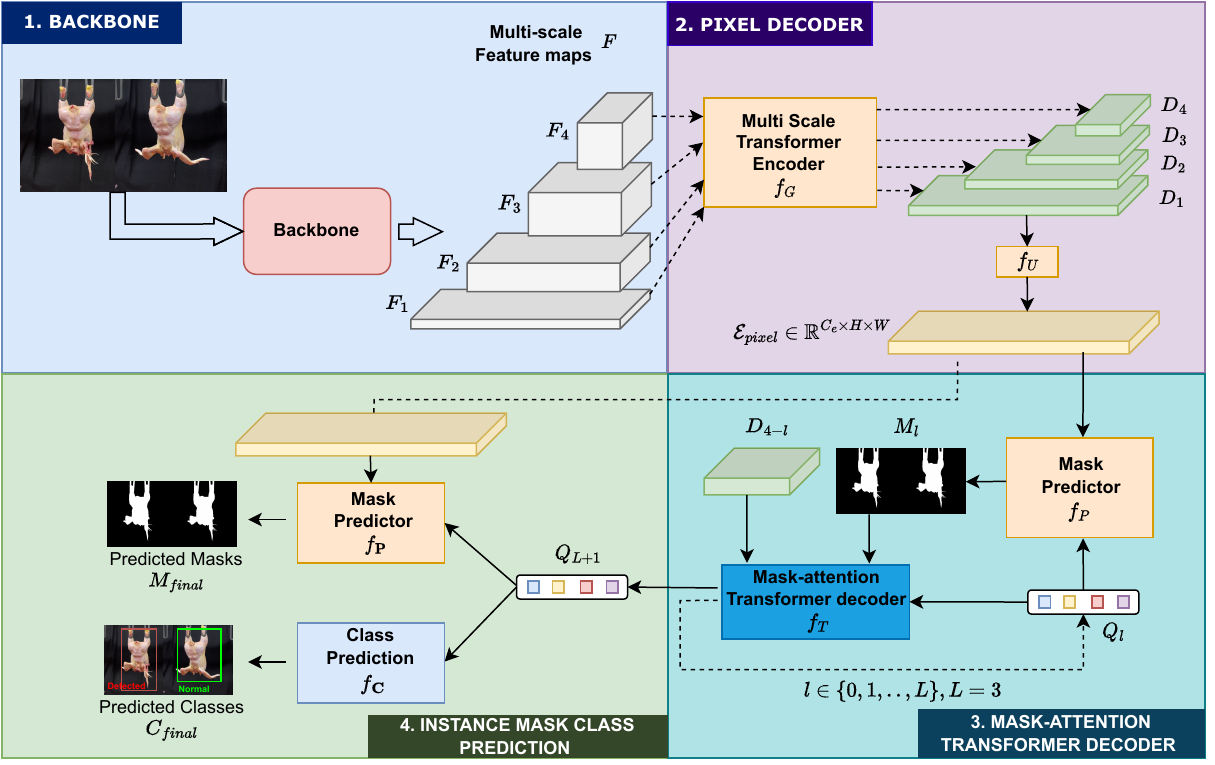}
\end{graphicalabstract}

\begin{highlights}
\item  \textbf{Dataset}: A poultry carcass dataset was acquired, comprising a total of 7,321 images gathered from real-world environments and collected from diverse chicken ages, chicken size, and number of chickens per frame. The dataset has been carefully annotated by three experts.
\item \textbf{Methodology}: CarcassFormer, an effective end-to-end Transformer-based framework, was proposed for simultaneously localizing poultry carcass regions, segmenting carcass areas, and determining carcasses with imperfections. CarcassFormer is based on Transformer-based  Unet architecture. \\
Our CarcassFormer is designed with four different components: Network Backbone to extract visual features, Pixel Decoder to utilize feature maps from various scales, Mask-Attention Transformer Decoder to predict the segmented masks of all instances, and Instance Mask and Class Prediction to provide segmentation mask and corresponding label of an individual instance. The extensive experiments showed that CarcassFormer outperforms both CNN-based networks, namely Mask R-CNN \cite{he2017mask} and HTC \cite{chen2019hybrid}, and Transformer-based networks, namely Mask2Former \cite{cheng2022masked} and QueryInst \cite{fang2021instances} on different backbone networks of ResNet-34 and ResNet-50 on various metrics of AP, AP@50, AP@75.
\item \textbf{Pre-trained models and Code}: The pre-trained model and source code of CarcassFormer is available for research purposes at: \url{https://github.com/UARK-AICV/CarcassFormer}.
\end{highlights}


\end{frontmatter}


\section{Introduction}
Increased consumption of poultry products will be a certainty for global food security achievement in the upcoming 30 years based on the efficiency of the utilization of poultry, as well as diverse consumer acceptance. The Food and Agriculture Organization of the United Nations 2005/2007 has projected that production of poultry will increase more than 100 percent by the year 2050 with an increased tonnage of poultry products, primarily broiler chickens, surpassing 180 million tons, with current projection estimated at just over 80 million tons \cite{alexandratos2012world}. Numerous studies have demonstrated increasing annual poultry consummation rates, mainly due to relatively inexpensive price, nutritional value, and health benefits \cite{elam2022live}. In the U.S., broiler chicken efficiency of feed utilization has increased 7 percent from 2021 to the present at a similar slaughter age between 47 and 48 days across the decade \cite{National}. With annualized increases in broiler production, concomitant increases in labor are necessary for meat production supply chain efficiency. In addition to the costs of increased workforce labor and workforce development, many poultry companies are suffering from labor shortages \cite{wu2022information} \cite{kaminski2020re}. Another negative side of relying on people for the process of poultry processing represents the varying results of carcass evaluation consistency. Thus, many companies use assembly lines stationed by employees to inspect the quality of chicken carcasses, which leaves room for human error and can result in miscategorized carcass defections. As a result, numerous agriculture industries, including poultry production facilities and poultry processing plant factories, are researching and investing in automated robotic technologies to improve processing and labor wellbeing, as well as profit \cite{ahlin2022robotic} \cite{ren2020agricultural} \cite{park2022artificial}. Further, there are numerous automation technologies offering noticeable economic benefits to agricultural production as of late \cite{jin2021development}.  

In the era of precision agriculture, Machine Learning (ML) and Computer Vision (CV) have emerged as high-performance computing technologies that are creating new opportunities to improve broiler management, production, and identification of processing defects with non-invasive low-cost techniques \cite{Aydin2017} \cite{Caldas-Cueva2021}.  In this study, the focus was on utilizing modern ML\&CV, i.e. Deep Learning, to analyze chicken carcasses after scalding, picking, and removal of head and feet in processing plants. Visual inspection is one of the most basic but also most important steps in controlling meat quality before the product is prepared, packaged, and distributed to the market. The image processing and classification within the poultry processing plants can optimize such systems, in addition to heightening food safety. Hence, our proposed intelligent and automated system will analyze and improve poultry processing concomitantly with increased data acquisition. Our computer vision system functions as an automated detection model capable of classifying defects and contaminated carcasses. While detection, segmentation, and classification are widespread tasks in computer vision \cite{dong2021survey, zhou2021deep, le2022deep}, they have focused on various tasks such as autonomous driving \cite{le2017deepsafedrive, le2017robust, janai2020computer, tong2020recent, truong2022otadapt, nguyen2022multi}, surveillance \cite{wray2021semantic, gabeur2020multi, yamazaki2022vlcap, vo2022aoe}, biometrics \cite{le2016novel, le2017semi, duong2019learning, duong2019mobiface, quach2022non}, and medical imaging \cite{han2017automatic, le2018deep, 9956588, tran2022ss, le2023scl, thang2023ai, nguyen2023embryosformer}, amodal understanding \cite{tran2022aisformer} which mainly target humans, car, objects, face, human organs. \emph{None of them target analyzing poultry carcass condemnations defects}. One of the main reasons is the lack of publicly available data.

In the context of poultry carcass analysis, distinguishing between single and multiple carcasses in an image is a crucial step for accurate quality assessment. To achieve this, the problem was approached as an instance segmentation task, involving the localization of individual instances. Additionally, mask classification was performed to determine whether a single poultry carcass was defective or not. While per-pixel classification (e.g FCN \cite{long2015fully}, Unet-based approaches \cite{ronneberger2015u, zhou2018unet++, ibtehaz2020multiresunet, le2021narrow, 9956588}) applies a classification loss to each output pixel and partitions an image into regions of different classes, mask classification (e.g Mask-RCNN \cite{he2017mask}, DETR \cite{zhu2020deformable}) predicts a set of binary masks, each associated with a single class prediction. 
\revise{
In recent years, there has been a significant growth in the adoption of Transformer architecture~\cite{vaswani2017attention} for semantic segmentation tasks. This trend is underscored by numerous approaches that have leveraged Transformer models, demonstrating state-of-the-art performance in the field. Notable examples include DETR~\cite{carion2020end}, SegFormer~\cite{xie2021segformer}, Mask2Former~\cite{cheng2022masked}, FASeg~\cite{he2023dynamic}, and Mask DINO~\cite{li2023mask}.
}
In this paper, the question of how to \emph{simultaneously handle both mask classification and pixel-level classification} is addressed.

To address the aforementioned question, we particularly leverage the recent Transformer technique \cite{vaswani2017attention} and propose \emph{CarcassFormer}, which aims to simultaneously localize poultry carcasses from moving shackles, segment the poultry carcass body, and classify defects or contaminated carcasses. To develop CarcassFormer, an experiment was set up at the University of Arkansas-Agricultural Experiment Station Pilot Processing Plant on the poultry research farm by placing a camera adjacent to the shackles of carcasses moving along a processing line. Each poultry carcass in the view of the camera will be analyzed by localizing with a bounding box, segmenting the boundary, and classifying to determine its imperfections. Any unapproved birds are then reworked. Notably, a bird is considered to be defective if it has one of the following issues: feathers, un-clean/dirty, skin peel, broken wings, or broken legs. The annotation requirement is following instructions provided by USAD \cite{Poultry-Grading-manual}.

Our contribution is three-fold as follows:
\begin{itemize}
    \item \textbf{Dataset}: A dataset containing a total of 7,321 images of poultry carcasses on a Pilot processing plant. The images in this diverse dataset contain real-world examples of chickens of a range of ages, sizes, and numbers of chickens per frame. The dataset has been carefully annotated by three experts. 
    
    \item \textbf{Methodology}: We propose CarcassFormer, an effective end-to-end Transformer-based framework for simultaneously localizing poultry carcass regions, segmenting carcass areas, and determining carcasses with imperfections. CarcassFormer is based on Transformer-based Unet architecture.

    \item \textbf{Pre-trained models and Code}: We will release our pre-trained model and source code of CarcassFormer for research purposes.
\end{itemize}

\subsection{Related Work}
\subsubsection{Image Segmentation}

Image segmentation is a critical computer vision task that involves dividing an image into different regions based on visual features. This process can be accomplished through either \emph{semantic segmentation} or \emph{instance segmentation}. Semantic segmentation categorizes pixels into multiple classes, e.g. foreground and background, but does not differentiate between different object instances of the same class. Popular semantic segmentation models include the Fully Convolutional Network (FCN) \cite{long2015fully} and its variants, such as the U-Net family \cite{ronneberger2015u, zhou2018unet++, ibtehaz2020multiresunet, le2021narrow}, as well as the Pyramid Scene Parsing Network (PSPNet) \cite{zhao2017pyramid} and DeepLabV3 \cite{chen2018encoder}. 

In contrast, instance segmentation aims to detect and segment individual objects by providing a unique segmentation mask for each object. There are two types of instance segmentation approaches: two-stage and one-stage methods. Two-stage approaches, such as top-down \cite{cai2018cascade, chen2019hybrid, cheng2020boundary} and bottom-up methods \cite{arnab2016bottom, chen2017deeplab, newell2017associative}, detect bounding boxes first and then perform segmentation within each region of interest. On the other hand, one-stage approaches, such as anchor-based methods \cite{li2017fully, bolya2019yolact} and anchor-free methods \cite{ying2019embedmask, chen2020blendmask, lee2020centermask}, perform both detection and segmentation simultaneously, resulting in less time consumption. Anchor-based one-stage approaches generate class-agnostic candidate masks on candidate regions and extract instances from a semantic branch. However, these approaches rely heavily on predefined anchors, which are sensitive to hyper-parameters. To address this issue, anchor-free one-stage methods eliminate anchor boxes and use corner/center points instead. Moreover, based on their feature backbone and learning mechanism, various approaches to instance segmentation can be categorized into either  Convolution Neural Network (CNN)-based or Transformer-based approaches as follows.

\subsubsection{CNN-based instance segmentation}
The idea of “detect then segment” has dominated in instance segmentation task, which is a two-stage method. In particular, Mask R-CNN \cite{he2017mask} is the most representative work. Based on the priority of detection and segmentation, there are two groups in this category: top-down methods and bottom-up methods. The former first predicts a bounding box for each object and then generates an instance mask within each bounding box \cite{he2017mask, o2015learning}. On the other hand, the latter associates pixel-level projection with each object instance and adopts a post-processing procedure to distinguish each instance \cite{arnab2016bottom, kong2018recurrent}. While the top-down methods mainly rely on the detection results and are prone to systematic artifacts on an overlapping instance, the bottom-up methods depend on the performances of post-processing and tend to suffer from under-segment or over-segment problems \cite{fathi2017semantic}. With a large amount of pixel-wise mask annotations, fully-supervised learning instance segmentation methods have achieved great performance. However, pixel-wise mask annotating is labor intensive (e.g., 22 hours to label 1000 segmentation masks \cite{lin2014microsoft}). Thus, weakly-supervised \cite{zhou2018brief, zhu2016weakly} and semi-supervised \cite{van2020survey} have been proposed. CNN-based image segmentation has been outreached in multiple Computer Vision tasks including amodal segmentation \cite{li2016amodal}, salient detection \cite{fan2019s4net}, human segmentation \cite{zhang2019pose2seg}, soft biometrics \cite{luu2016deep}. CNN-based instance segmentation survey can be found at \cite{hafiz2020survey, gu2022review}.

\subsubsection{Transformer in Computer Vision}
Transformer was first introduced by \cite{vaswani2017attention} for language translation and obtained State-Of-The-Art (SOTA) results in many other language processing tasks. Recently, many models \cite{carion2020end}, \cite{liu2022dab}, \cite{li2022dn} successfully applied the Transformer concept to computer vision and achieved promising performance. The core idea behind transformer architecture \cite{vaswani2017attention} is the self-attention mechanism to capture long-range relationships. It has obtained state-of-the-art in many Natural Language Processing (NLP) tasks. Besides, Transformers have worked well suited for parallelization, facilitating training on large datasets
Transformer has been successfully applied to enrich global information in various tasks in Computer Vision such as image recognition \cite{dosovitskiy2020image, touvron2021training} object detection \cite{carion2020end, zhu2020deformable, sun2021sparse}, image segmentation \cite{ye2019cross, zheng2021rethinking, tran2022aisformer}, action localization \cite{vo2021aei, vo2022aoe}, video captioning \cite{ yamazaki2022vlcap, yamazaki2022vltint}.
DETR \cite{zhu2020deformable} is the first model that uses Transformer as an end-to-end and query-based object detector, with bipartite-matching loss and set prediction objective. Inspired by \cite{zhu2020deformable, cheng2021per}, which are end-to-end prediction objectives and successfully address multiple tasks without modifying the architecture, loss, or the training procedure, the merits of Transformer were inherited and CarcassFormer was proposed. Our network is an end-to-end Transformer-based framework and simultaneously tackles both segmentation and classification tasks.   

\revise{
Transformer-based networks have also found application in addressing detection and segmentation challenges within poultry science.  ~\cite{lin2022judgment} proposes a vision transformer model to screen the breeding performance of roosters by analyzing correlations between cockscomb characteristics and semen quality, aiming to overcome the time-consuming and error-prone nature of human-based screening. ~\cite{hu2023attention} improves pig segmentation in farming environments using a grouped transformer attention module with Mask R-CNN networks and data augmentation. ~\cite{zhao2023research} proposes a real-time mutton multi-part classification and detection method using Swin-Transformer. ~\cite{he2023reliable} presents Residual-Transformer-Fine-Grained (ResTFG), a model merging transformer and CNN for precise classification of seven chicken Eimeria species from microscopic images.
}

\begin{figure*}[!b] 
    \centering
    \includegraphics[width=1.\linewidth]{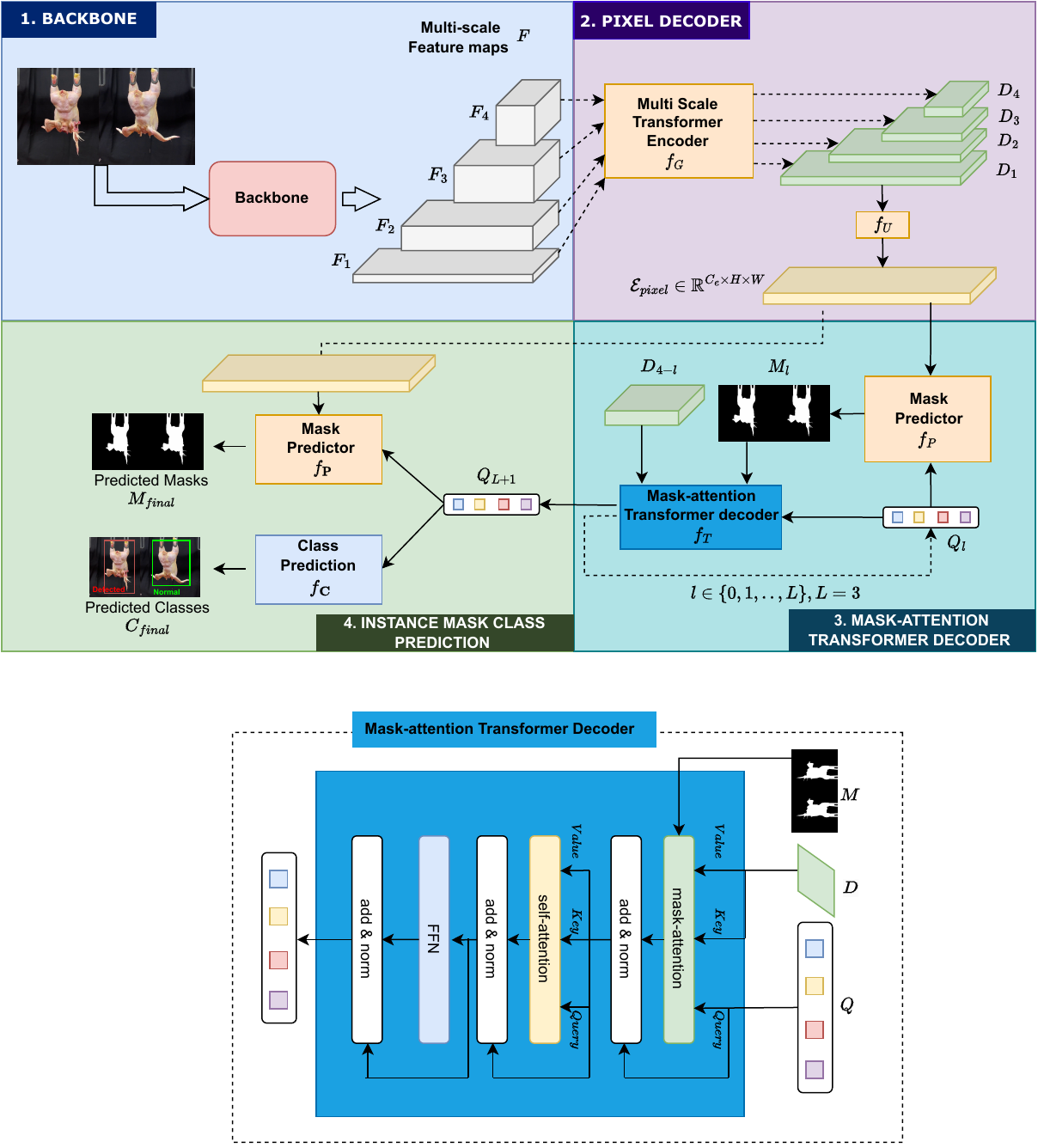}
    \caption{ Top: Overall flowchart of our proposed CarcassFormer consisting of four components: 1. Network Backbone; 2. Pixel Decoder; 3. Mask-Attention Transformer Decoder; 4. Instance Mask Class Prediction. Bottom: Details of third component Mask-Attention Transformer Decoder.}
    \label{fig:model}
\end{figure*}

\section{Materials and  Methods}

\subsection{Data Collection }
\label{sec:data}
The data was collected at the University of Arkansas pilot processing  plant (Fayetteville, AR). Multiple broiler chicken products at different ages were processed using standard commercial practices and following rigorous  animal-handling  procedures  that  are  in  compliance  with federal and institutional regulations regarding proper animal care practices \cite{FASS2010}. The video-capturing system was set up in the area after feather picking and before chilling and evisceration. We decided on his system placement so that three common kinds of defects that can occur during normal processing could be evaluated, namely tearing of the skin, presence of feathers, and broken/disjointed bones.

\begin{figure*}[!t]
\begin{minipage}[t]{0.5\textwidth}
    \centering \includegraphics[width=0.9\linewidth]{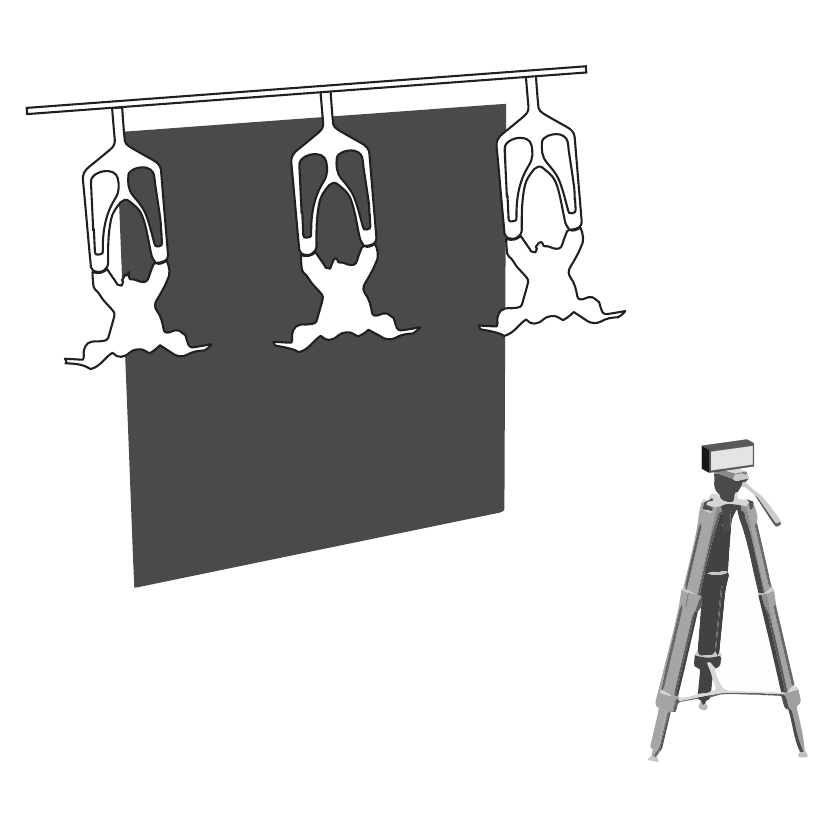}
    \vspace{-0.5em}
    \caption{\textbf{Camera setup for data collection.} A black curtain is hung behind the shackle to provide a certain contrast to the carcasses. A camera is placed to capture the carcasses within the black curtain.}
    \label{fig:camera-setup}
\end{minipage} 
\hspace{2.mm}
\begin{minipage}[t]{0.5\linewidth}
    \centering
    \includegraphics[width=0.8\linewidth]{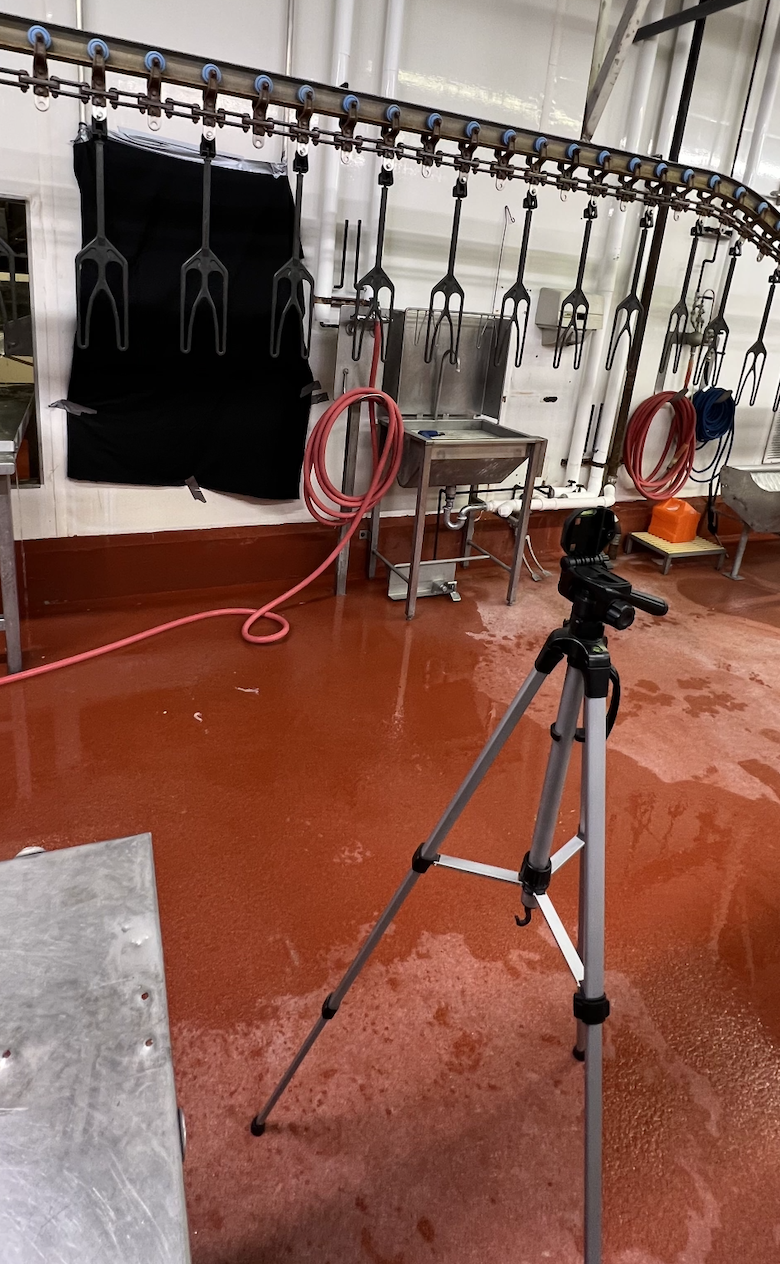}
    \caption{\revise{An overview image of the shooting location. The black curtain is hung on the wall behind the shackle.}}
    \label{fig:location_setup}
\end{minipage}
\end{figure*}

\begin{figure*}[!h] 
    \centering
    \includegraphics[width=0.9\linewidth]{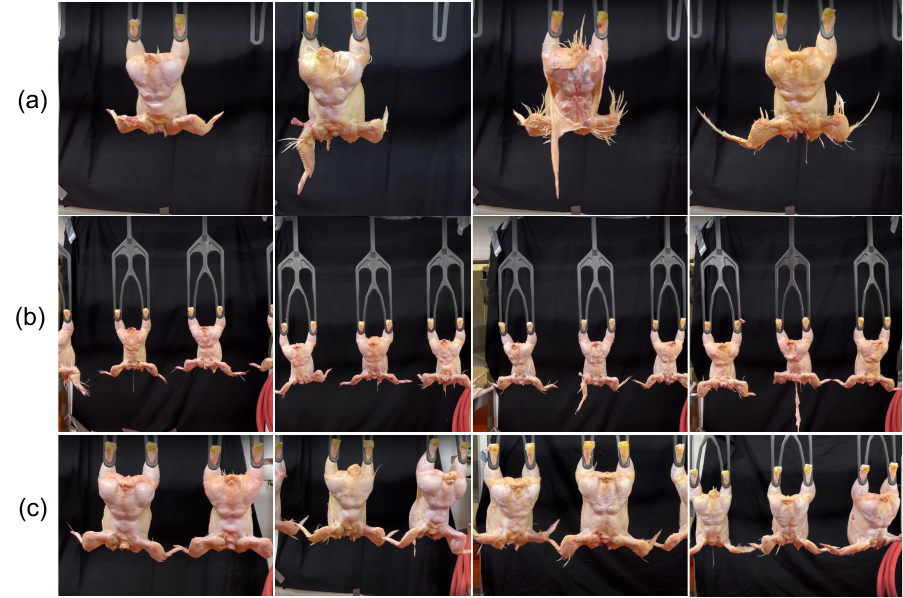}
    \caption{Illustrations of \textbf{data collected}, which comprises (a) single carcass/instance per image/frame; (b) multiple carcass/instance per image/frame; (c )carcass/instance at different scale/resolution. The carcass is processed with various defects such as tearing of skin, feathers, broken/disjointed bones.}
    \label{fig:data_illustration}
\end{figure*}

To obtain the dataset named CarcassDefect, a camera was set up in front of the shackle line, whereas a black curtain was hung behind the shackle. Videos were recorded at 10 frames per second. The camera setup can be visualized in Fig. \ref{fig:camera-setup} \revise{and Fig. \ref{fig:location_setup}.} 
 In the end, a total of 7,321 images were collected, comprising 4,279 single carcass images and 3,042 multiple carcass images. Fig.\ref{fig:data_illustration} illustrates some images from our CarcassDefect dataset, which comprised a large diversity of carcass quality such as resolution (small carcass, large carcass), the number of carcass per image (a single carcass per image, multiple carcasses per image), various defect (carcass defect can be the one with tearing of skin, feathers, broken/disjointed bones.), etc.


\begin{table}[!h]
\caption{The distribution of the \textbf{images} in CarcassDefect Dataset in regard to the normal carcass and defective carcass at both a single carcass per frame and multiple carcasses per frame. }
\renewcommand{\arraystretch}{1.3}
\setlength{\tabcolsep}{4pt}
\resizebox{\textwidth}{!}{
\begin{tabular}{l|c|c|c|c}
\toprule
 & \textbf{Single carcass per image}  & \textbf{Multiple carcasses per image} & \textbf{Total} \\ \toprule
\textbf{Trainset}   & 3,017 & 2,115 &   5,132    \\ \hline
\textbf{Valset}    & 754 & 535 &    1,289               \\ \hline
\textbf{Testset}    & 508 & 392 &    900               \\ \hline
\textbf{Total}     & 4,279 &  3,042 &    7,321               \\ 
\bottomrule
\end{tabular}}
\label{tab:img}
\end{table}

\begin{figure*}[!h] 
    \centering
    \includegraphics[width=0.95\linewidth]{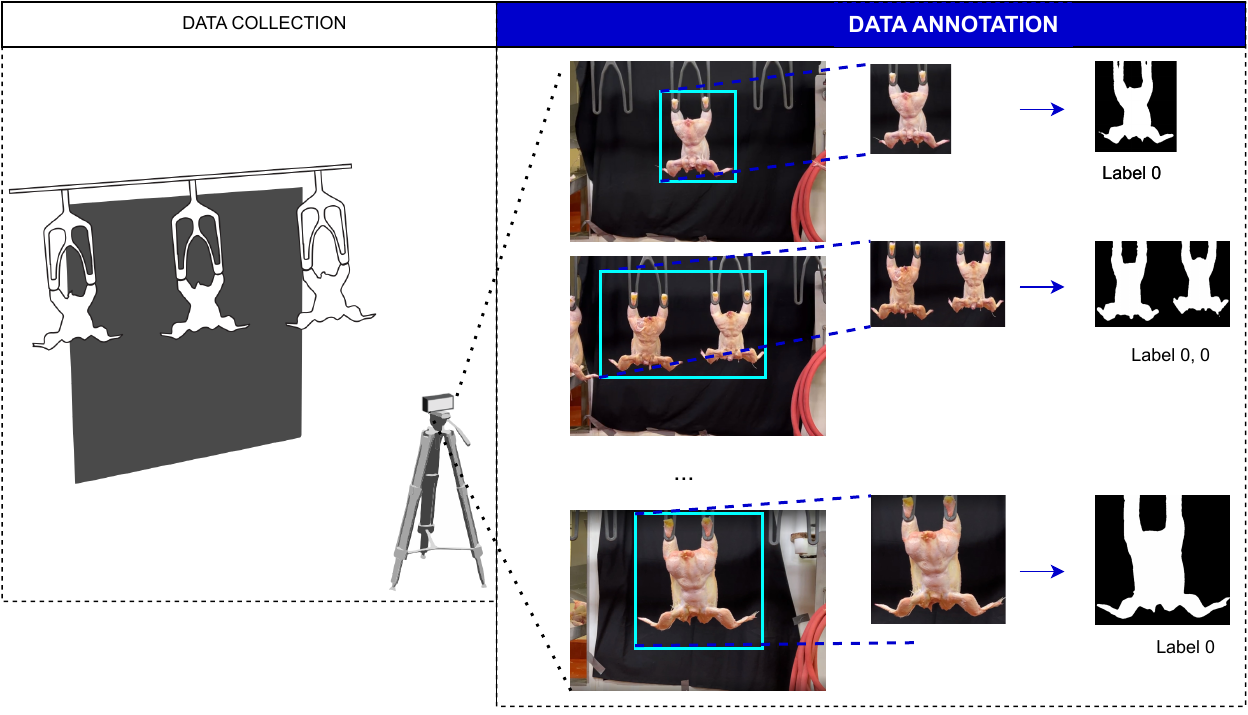}
    \caption{An illustration of \textbf{Data Annotation Process}. Each frame from the recorded video is annotated with bounding boxes for detection,  masks for segmentation, and defect labels for classification.}
    \label{fig:dataannotation}
\end{figure*}

\subsection{Data Annotation}
\begin{figure}[!h] 
    \centering
    \includegraphics[width=0.75\linewidth]{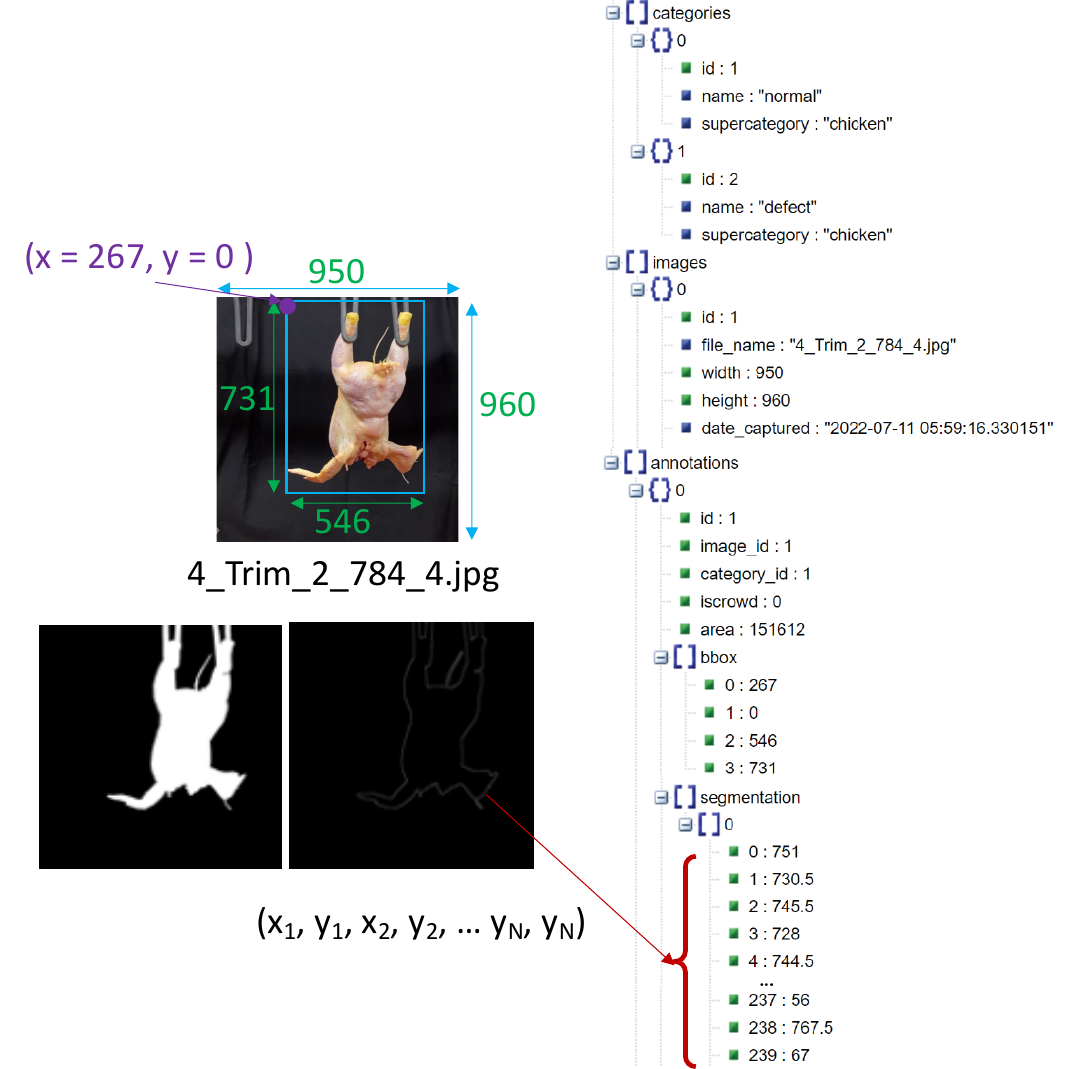}
    \caption{A demonstration of data annotation in JSON file following COCO format. }
    \label{fig:data_coco}
\end{figure}

Upon acquiring the video data, the next crucial step is to annotate the images extracted from the footage to generate training data for detection, segmentation, and classification tasks. Annotation involves labeling each frame with bounding boxes for detection, masks for segmentation, and labels for classification. This process enables the computer vision system to learn from the annotated data, which enhances its ability to perform these tasks accurately and efficiently. The data annotation process is illustrated in Fig.\ref{fig:dataannotation}. The annotated data is saved in a JSON file and follow the COCO format \cite{lin2014microsoft} as demonstrated in Fig.\ref{fig:data_coco}. In this COCO format, the data is described as follows:
\begin{itemize}
    \item \textit{categories}:  defined as `normal' and `defect' presenting labeled in the dataset. The defect class is determined for a carcass that has either `feather' or `broken wings' `broken legs' or `peeled skin'.
    \item \textit{images}: frames extracted from recorded videos.  \textit{images} is a list of objects with meta-data information about the image. An object includes the following keys:
    \begin{itemize}
        \item \textit{id}: a unique identifier that differentiates each image within a list. It can be defined as the file name.
        \item \textit{file\_name}: the name of the file. In the example (Fig.\ref{fig:data_coco}).
        \item \textit{width}: the image height such as 950  pixels. 
        \item \textit{height}: the image height such as 960  pixels.
        \item \textit{date\_captured}: the date and time when the image was captured.
    \end{itemize}
    \item \textit{annotations}: contain all meta-data about the labels related to an object. They are a bounding box, segmentation mask and classification label.
    \begin{itemize}
        \item \textit{id}: The index of instance.
        \item \textit{image\_id}: The index of the corresponding image. This \textit{image\_id} is corresponding to \textit{id} in \textit{images}.
        \item \textit{category\_id}: This is category id which is defined in \textit{categories}. In our case, \textit{category\_id} is either `1' - normal or `0' - defect.
        \item \textit{iscrowd}: if there are multiple instances/carcasses in the image, \textit{iscrowd} is set as 1. Otherwise, \textit{iscrowd} is set as 0 if there is a single instance/carcass in the image.
        \item \textit{area}: is the area of instance in the image.
        \item \textit{bbox}: The bounding box determines an object’s location represented as [xmin, ymin, width, height] where the (xmin, ymin) coordinates correspond to the top-left position of an object and (width, height) are width and height of the object. In the example shown in Fig. \ref{fig:data_coco}, xmin = 27, ymin = 0, width = 546, height = 731.
        \item \textit{segmentation}: The segmentation mask is specified by Run-length encoded (RLE) values \cite{golomb1966run}.
    
    \end{itemize}
\end{itemize}

The data statistic of our CarcassDefect dataset is shown in Table \ref{tab:img} and Table \ref{tab:ins}. Table \ref{tab:img} shows the distribution of the image between a single carcass per image and multiple carcasses per image between two categories of normal and defect. Table \ref{tab:ins} shows the distribution of the instance between a single carcass per image and multiple carcasses per image between two categories of normal and defect.

\begin{table}[!h]
\caption{The distribution of the \textbf{instances} in CarcassDefect Dataset regarded as normal and defective carcasses at both a single carcass per frame and multiple carcasses per frame. }
\renewcommand{\arraystretch}{1.3}
\setlength{\tabcolsep}{4pt}
\resizebox{\textwidth}{!}{
\begin{tabular}{l|cc|cc|l}
\toprule
\multirow{2}{*}{} & \multicolumn{2}{l|}{\textbf{Single carcass per image} }  & \multicolumn{2}{l|}{\textbf{Multiple carcasses per image} } & \multirow{2}{*}{\textbf{Total}} \\ \cline{2-5}
                        &  {\textbf{Normal}} & \textbf{Defeat} &  {\textbf{Normal}}  & \textbf{Defeat} &                        \\ \toprule
\textbf{Trainset}   &  {1,302}  & 1,715  &  {1,571}   & 1,842  & 6,430                  \\ \hline
\textbf{Valset}                  &  355    & 399    &  {422}     & 466    & 1,642                  \\ \hline
\textbf{Testset}                 &  320    & 188    &  {359}     & 267    & 1,134                  \\ \hline
\textbf{Total}                   &  {1,977}  & 2,302  &  {2,352}   & 2,575  & 9,206 \\
\bottomrule
\end{tabular}
}
\label{tab:ins}
\end{table}

\subsection{Proposed method}
In the sections below, the proposed end-to-end transformer-based framework, termed CarcassFormer, is introduced for chicken carcass detection, segmentation, and carcass defect classification. Figure \ref{fig:model} illustrates the flowchart of our CarcassFormer network consisting of four key parts: Backbone, Pixel Decoder, Multi-Scale Transformer Encoder, and Masked-attention Transformer Decoder.
To train CarcassFormer, the  stochastic gradient descent (SGD) optimizer was utilized with a learning rate of 0.0001 and a batch size of 4 over 100 epochs. The experiments were conducted using an Intel(R) Core(TM) i9-10980XE 3.00GHz CPU and a Quadro RTX 8000 GPU.

\subsection{Backbone}
A backbone network, a foundational architecture employed for feature extraction, is typically pre-trained on a variety of tasks and has demonstrated its effectiveness across various domains. AlexNet \cite{krizhevsky2017imagenet} is regarded as the inaugural Deep Learning (DL) backbone. The VGG family, which includes VGG-16 and VGG-19 \cite{simonyan2014very}, is one of the most prevalent backbones utilized in computer science endeavors. In contrast to AlexNet and VGG, ResNets \cite{he2016deep} are based on Convolutional Neural Networks (CNNs) and were developed concomitantly with the introduction of residual networks. ResNet variants, such as ResNet-18, ResNet-34, ResNet-50, ResNet-101, and ResNet-151, are extensively employed for object detection and semantic segmentation tasks. Following the advent of ResNets, numerous other CNN-based backbones have been proposed, including Inception \cite{szegedy2015going}, DenseNet \cite{huang2017densely}, DarkNet \cite{lin2013network}, ShuffleNet \cite{zhang2018shufflenet}, MobileNet \cite{howard2017mobilenets}, and YOLO \cite{huang2018yolo}. 
\revise{Recently, there has been a significant advancement in backbone architectures, incorporating transformer architecture~\cite{vaswani2017attention}, along with leveraging the multi-scale features of ResNet. Prominent examples of these advancements include ViT~\cite{dosovitskiy2020image}, PvT~\cite{zhao2017pyramid}, and Swin~\cite{liu2021swin}.}
In the present work, ResNet (i.e., ResNet-18, 34, 50) \cite{he2016deep} \revise{and Swin (i.e., Swin-T)} are employed as the backbone network.

By utilizing ResNets \cite{he2016deep} as the backbone network, an input image $I$ with dimensions $H \times W$ is transformed into a multi-scale feature $F$, specifically a set of four feature maps ${F_i}{i=1}^4$. These feature maps are represented as $F_1 \in \mathbb{R}^{C{F_1}\times \frac{H}{4} \times \frac{W}{4}}$, $ F_2 \in \mathbb{R}^{C_{F_2}\times \frac{H}{8} \times \frac{W}{8}}$, $ F_3 \in \mathbb{R}^{C_{F_3}\times \frac{H}{16} \times \frac{W}{16}}$, and $ F_4 \in \mathbb{R}^{C_{F_4}\times \frac{H}{32} \times \frac{W}{32}}$, where $C_{F_1}, C_{F_2}, C_{F_3}, C_{F_4}$ denote the number of channels.

\subsection{Pixel Decoder}
This module enhances the multi-scale features of an image by utilizing four feature maps from the backbone. It consists of two parts: the Multi-Scale Transformer Encoder (section \ref{multiscale_tf_encoder}) and the Per-pixel Embeddings Module (section \ref{perpixel}). In general, the Multi-Scale Transformer Encoder uses an attention mechanism to learn the correlation between the multi-scale feature maps $F_1, F_2, F_3, F_4$. This results in corresponding, richer encoded feature maps $D_1, D_2, D_3, D_4$. Meanwhile, the per-pixel embeddings Module takes the encoded feature map ($D_1 
  $) to compute the per-pixel embeddings $\mathcal{E}_{pixel}$  of the image.

\subsubsection{Multi Scale Transformer Encoder}
\label{multiscale_tf_encoder}

This module takes the last three features from the backbone, ordered from low to high resolution (i.e., $F_4, F_3$, $F_2$, and $F_1$), are processed in a hierarchical fashion. 
These three features first go through an embedding projection $f_E$ to achieve the flattened embed feature $S_i$ with a consistent channel size $C_e$. Note that the value of $C_e$ is equal to $C_{F_{1}}$, and this is specifically intended for computing the per-pixel embeddings in section \ref{perpixel}.
\begin{equation}
    S_i = f_E (F_i) 
\end{equation}
where $i = \{4,3,2,1\}$, $S_i \in \mathbb{R}^{H_i\cdot W_i\times C_e}$, $f_E$ is a $1 \times 1$ convolution layer with the output dimension of $C_e$, followed by a flatten layer. The purpose of the flatten layer is to prepare $S_i$ as input for a transformer layer, which requires a sequence of embedding features rather than spatial features. 

To investigate the correlated feature embeddings between different levels, the flattened embedding features $S_i$ from each multi-scale level were concatenated and passed through a transformer encoder. This involves merging the flattened embeddings from different levels into a single input sequence for the transformer encoder.
\begin{equation}
S = [S_i]_{i\in\{4,3,2,1\}}    
\end{equation}
where $S \in \mathbb{R}^{ K \times C_e}$, $K$ is the total number of embedding feature, $K = \sum_{i\in\{4,3,2,1\}} H_i \cdot W_i $. 

However, since the current embedding features $S$ are flattened out of their original spatial shapes and concatenated from multiple levels, they do not include information about the spatial location and scale level of each feature. To address this issue, each embedding feature in $S$ is supplemented with two types of learnable encoding. The first is a positional encoding that provides spatial information about the original location of each feature within the image. The second is a level encoding that enables the transformer encoder to distinguish between features from different scales. By incorporating these encodings, spatial and scale level information is preserved during the calculation process. Let denote the learnable positional encodings as $P$ and the learnable level encodings as $L$, where $P$ and $L$ share the same shape with $S$, $P, L \in \mathbb{R}^{K\times C_e}$.

Let denote the Multi Scale Transformer Encoder as $f_G$, this Transformer follows the architecture designed in \cite{dosovitskiy2020image}. This transformer encoder produces learned features from the input sequence. It takes a sequence of embedded features and outputs encoded features that capture the relationships between the elements. These encoded features retain important information while removing redundancy. The encoder computes each encoded feature using a self-attention mechanism \cite{vaswani2017attention}, allowing it to selectively focus on the most relevant features and capture long-range dependencies, making it effective for enriching multiscale features. Formulary, the correlated feature embeddings between different levels $E \in \mathbb{R}^{K\times C_e} $ is computed by passing $S, L, $ and $P$ through $f_G$.
\begin{equation}
    E = f_G (S, P, L)
\end{equation}

The correlated feature embedding $E$ is divided into groups based on the multi-scale level, denoted as $E_i$ where $i\in \{4,3,2,1\}$ and $E_i \in \mathbb{R}^{H_i\cdot W_i\times C_e}$. Next, each $E_i$ is restored to its original spatial shape by unflattening, resulting in an output enriched multi-scale feature map $D_i$.
\begin{equation}
    D_i = \text{unflatten} (E_i)
\end{equation}
where $D_i \in R^{C_e\times H_i\times W_i  }$. 

In summary, this module takes the multi-scale feature map $ F_4 \in \mathbb{R}^{C_{F_4}\times \frac{H}{32} \times \frac{W}{32}}$ , $ F_3 \in \mathbb{R}^{C_{F_3}\times \frac{H}{16} \times \frac{W}{16}}$, $ F_2 \in \mathbb{R}^{C_{F_2}\times \frac{H}{8} \times \frac{W}{8}}$ and $F_1 \in \mathbb{R}^{C{F_1}\times \frac{H}{4} \times \frac{W}{4}}$ and outputs the enriched multi-scale feature map $ D_4 \in \mathbb{R}^{C_e\times \frac{H}{32} \times \frac{W}{32}}$, $ D_3 \in \mathbb{R}^{C_e\times \frac{H}{16} \times \frac{W}{16}}$, $ D_2 \in \mathbb{R}^{C_e\times \frac{H}{8} \times \frac{W}{8}}$ and $ D_1 \in \mathbb{R}^{C_e\times \frac{H}{4} \times \frac{W}{4}}$, which captures the correlation and important information while removing redundancy.

\subsubsection{ Per-pixel Embeddings Module} 
\label{perpixel}
This section describes the second stage of Pixel Decoder, where the per-pixel embedding $\mathcal{E}_{pixel}$ is computed.
This module takes the encoded feature map $D_1 \in \mathbb{R}^{C_e\times \frac{H}{4} \times \frac{W}{4}}$ from the Multi Scale Transformer Encoder module as input.
The per-pixel embedding $\mathcal{E}_{pixel}$ is computed as follow:
\begin{equation}
\begin{split}
    \mathcal{E}_{pixel} &= f_U (D_1)
\end{split}
\end{equation}
The function $f_U$ is a sequence of two $2\times2$ transposed convolutional layers with stride $2$, which scales up the spatial shape of $D_1$ four times from $\frac{H}{4} \times \frac{W}{4}$ to the original image's spatial shape of $H\times W$. As a result, $\mathcal{E}_{pixel}$ has a dimension of $\mathbb{R}^{C{e}\times H\times W }$.

Intuitively, each pixel feature of $\mathcal{E}_{pixel}$ represents both the semantic and the mask classification feature of the corresponding pixel on the original image.

\subsection{Mask-attention Transformer Decoder}
\subsubsection{Mask Predictor}
To predict the segmented masks of possible instances in an image, per-pixel embeddings $\mathcal{E}_{pixel} \in \mathbb{R}^{C{e}\times H\times W }$ were utilized. These embeddings represent both the semantic and mask classification features of each corresponding pixel on the original image.

Then, the prediction process involves learning $N$ per-segment query embeddings $Q \in \mathbb{R}^{N\times C_e}$, which represent the features of the maximum $N$ possible instances in the image. 
Each instance query embedding correlates with every single pixel feature in $\mathcal{E}_{pixel}$ to determine whether the pixel belongs to the corresponding instance or not. Therefore, the predicted instance segmentation mask was derived as follows:
\begin{equation}
    M = f_P (Q, \mathcal{E}_{pixel})
\end{equation}

where $M \in \mathbb{R}^{N\times H\times W}$, which are $N$ masks of $N$ possible instances in the image. The Mask Predictor $f_P$ is a simple dot product on the feature channel $C_e$, followed by a sigmoid activation.

\subsubsection{Mask-attention Transformer Decoder}

The Mask-Attention Transformer Decoder $f_T$ was employed to obtain effective per-segment query embeddings $Q \in \mathbb{R}^{N \times C_e}$ that represented instances in the image. This decoder applies attention to the image features, allowing it to decode the per-segment query embeddings and capture the instance mask feature.

The third blob in our overall flowchart (Figure \ref{fig:model}) illustrates the procedure on applying the Mask-Attention Transformer Decoder. In general, this module decodes $N$ per-segment query embeddings $Q \in \mathbb{R}^{N \times C_q}$ from the encoded feature maps $D_1$, $D_2$, $D_3$, and $D_4$. These query embeddings represent the features of the maximum $N$ possible instances in the image.

The decoding procedure is performed recurrently, with each step treated as a layer (denoted as $l$) and beginning at $l = 0$. The encoded feature maps $D$ have four levels, denoted as $D_4$, $D_3$, $D_2$, and $D_1$. Therefore, this recurrent process occurs four times, progressing from the lowest to highest resolution encoded feature maps.  During each recurrent step, the encoded feature maps that are considered are $D_{4-l}$, where $l$ represents the current layer. This means that during the first recurrent step ($l = 0$), the encoded feature maps that are used are $D_4$. During the second recurrent step ($l = 1$), the encoded feature maps that are used are $D_3$. During the third recurrent step ($l = 2$), the encoded feature maps that are used are $D_2$. Finally, during the fourth recurrent step ($l = 3$), the encoded feature maps that are used are $D_1$. At each layer, the queries $Q_{l+1}$ are decoded from the previous layer's query $Q_l$ and the corresponding encoded feature maps. 

Additionally, a predicted mask, $M_l$ is computed by using the current query embeddings $Q_l$ and the per-pixel embeddings $\mathcal{E}_{pixel}$. The resulting mask is then interpolated to the same size as the current feature map $D_{4-l}$. This mask is used as an attention mechanism that helps the query embeddings to focus on the most salient parts of the feature maps. Specifically, during the decoding process, the attention mask is applied to the encoded feature maps $D_{4-l}$, allowing the query embeddings to selectively attend to certain regions of the feature maps that are most relevant to the instance being decoded. Formularly, at each recurrent step: 
\begin{equation}
\begin{split}
    M_l &=  f_{P} (Q_l, \mathcal{E}_{pixel}) \\
    Q_{l+1} &= f_{T} (Q_{l}, D_{4-l}, M_l)  
\end{split}
\end{equation}

\subsection{Instance Mask and Class Prediction}
The procedure can be visualized using the fourth component of the overall flowchart, as shown in Figure \ref{fig:model}. In this step, the query encoder $Q_L$ (where $L = 3$) and the per-pixel embeddings $\mathcal{E}{pixel}$ are utilized to compute the output instance segmentation masks, denoted as $M_{final}$. These masks, represented by a tensor $M_{final} \in \mathbb{R}^{N\times H \times W}$, correspond to $N$ possible instances within the image.

To generate the masks, the function $f_P$ takes $Q_L$ and $\mathcal{E}{pixel}$ as inputs:
\begin{equation}
M_{final} = f_P (Q_L, \mathcal{E}_{pixel})
\end{equation}

Moreover, alongside the masks, the semantic class of each instance is predicted using another function called $f_C$. This function is implemented as a Multi-Layer Perceptron (MLP) with two hidden layers. It takes the per-segment embeddings $Q_L$ as input and produces $N$ semantic classes, represented by $C_{final} \in \mathbb{R}^{N\times C}$. Here, $C$ represents the number of semantic categories.

The prediction of semantic classes can be expressed as:
\begin{equation}
C_{final} = f_C (Q_L)
\end{equation}

By combining these steps, this module is able to generate both instance segmentation masks ($M_{final}$) and predict the semantic class labels for each instance ($C_{final}$) using the decoded instance queries ($Q_L$) and per-pixel embeddings ($\mathcal{E}_{pixel}$).

\subsection{Metrics}
\label{sec:metrics}

We adopt Average Precision (AP) to evaluate the method.  AP quantifies how well the model is able to precisely locate and classify objects (e.g. defect or normal) within an image. The AP computation from MSCOCO \cite{lin2014microsoft} was followed.

In the recognition task, each image is associated with a single prediction for classification. Evaluating the model became straightforward as the accuracy metric could be calculated, measuring the ratio of correct predictions.
On the other hand, in the field of object detection and classification, a prediction comprises a bounding box or a segmentation mask that helps locate the object, along with the predicted category for that object. To determine a \textbf{correct prediction}, two criteria are considered. Firstly, the prediction must have an Intersection over Union (IoU) value greater than a threshold $\epsilon$ when compared to the actual box or mask of the object. Secondly, the prediction must accurately classify the category of the object.
In addition, for each image, a method can output multiple predictions and the number of predictions can be higher or lower than the actual object within the image. Thus, precision and recall metrics are taken into account. Precision is the ratio of correct predictions to the total number of predictions (Equation \ref{eq:precision}). Precision can be considered as a measure of how precise the model's predictions were in terms of correctly detecting objects.
\begin{equation}
    Pre = \frac{\text{number of correct predictions}}{\text{number of predictions}}
    \label{eq:precision}
\end{equation}
Meanwhile, recall is the ratio of correct predictions to the total number of actual objects within the image (Equation\ref{eq:recall}). It can be thought of as a measure of how comprehensive the model's predictions are in terms of capturing all the objects present.
\begin{equation}
    Rec = \frac{\text{number of correct predictions}}{\text{number of actual objects}}
    \label{eq:recall}
\end{equation}

Average Precision (AP) calculates the average precision across different recall values. Specifically, AP is computed at different IoU thresholds $\epsilon$, which determine what is considered a correct prediction. For instance, when the threshold $\epsilon$ is set to 50\%, it is denoted as AP@50. 
Let's consider an image with a list of actual ground truth objects denoted as $A = \{a_1, a_2, ..., a_n\}$, and a method that generates $m$ predictions denoted as $B = \{b_1, b_2, ..., b_m\}$. The predictions are sorted in descending order based on their confidence scores. In the process, the sorted list $B$ was iterated through, and at each step, the correctness of the prediction $b_i$ (where $i \in \{1, 2, ..., m\}$) was determined. This is done by checking if the category is correctly matched and if the IoU is greater than the specified threshold $\epsilon$. The number of correct predictions at this step was kept track of, denoted as $C_i$.

Using $C_i$, the precision $Pre_i$ and recall $Red_i$ could be computed at each step. The iteration stops when $Rec_i = 1$, indicating that all the objects have been captured, or when all the predictions were iterated through. The AP@$\epsilon$ is then computed as follows:
\begin{equation}
AP@\epsilon = \int_{0}^{1} Pre(Rec) , dRec
\end{equation}

The reported AP in our table is the average of AP values ranging from AP@50 to AP@95, with a step size of 5\% as depicted in Equation \ref{eq:ap}. This provides a comprehensive evaluation of the model's performance across different IoU thresholds and recall levels.

\begin{equation}
AP = \frac{1}{10}\sum_{\epsilon = 0.5 ;  \epsilon+=0.05}^{0.95}{AP@\epsilon}
\label{eq:ap}
\end{equation}

\revise{We also quantify the model's complexity using three key metrics: the number of floating-point operations (FLOPs), the count of model parameters (Params), and the frames processed per second (FPS). FLOPs are computed as an average over 100 testing images.
FPS is evaluated on a Quadro RTX 8000 GPU with a batch size of 1, calculated as the average runtime across the entire validation set, inclusive of post-processing time. 
}

\section{Results and Discussion}

\subsection{Implementation Details}

The implementation of the pixel decoder in this study involves the use of an advanced multi-scale deformable attention Transformer (DERT) as described in \cite{zhu2020deformable}. Specifically, the DERT is applied to feature maps with resolutions of 1/8, 1/16, and 1/32. A simple upsampling layer with a lateral connection is then employed on the final 1/8 feature map to generate the per-pixel embedding feature map of resolution 1/4. The Transformer encoder used in this study is configured with L=3 and a set of 100 queries.

\subsection{Quantitative Performance and Comparison}
\label{sec:performance}

In this section, our proposed CarcassFormer was evaluated on the Carcass dataset (Section \ref{sec:data}), which consisted of two subsets corresponding to a single carcass per image and multiple carcasses per image. The performance of CarcassFormer on various metrics, as shown in Table \ref{tab:detail_result_single} and Table \ref{tab:detail_result_single} for different tasks, was detailedly reported.
The authors then compared CarcassFormer with both CNN-based networks, namely Mask R-CNN \cite{he2017mask} and HTC \cite{chen2019hybrid}, as well as Transformer-based networks, namely \revise{Mask DINO~\cite{li2023mask}}, Mask2Former \cite{cheng2022masked} and QueryInst \cite{fang2021instances}. The comparison was conducted using \revise{three} different backbone networks: ResNet-34, ResNet-50, \revise{and Swin-T}. The performance for two subsets was reported: a single carcass per image (Table \ref{tab:single_r34}, Table \ref{tab:single_r50}) and multiple carcasses per image (Table \ref{tab:compare_overlap_r34}, Table \ref{tab:compare_overlap_r50}\revise{, Table \ref{tab:compare_overlap_swint}}). For each table, the metrics for detection, classification, segmentation, \revise{and model complexity} were reported, as defined in Section \ref{sec:metrics}.
\begin{table}[!h]
    \centering
    \caption{Detailed Performance of CarcassFormer on Single Carcass Per Image Dataset on both Detection and Segmentation, whereas $AP_{normal}$ \& $AP_{defect}$ include classification results.}
    \renewcommand{\arraystretch}{1.3}
    \setlength{\tabcolsep}{3pt}
    \resizebox{1.0\textwidth}{!}{
    \begin{tabular}{c|c|cccccc|ccc}
    \hline
        \textbf{Backbone} & \textbf{Task} & $AP$ & $AP_{50}$ & $AP_{75}$ & $AP_{95}$ & $AP_{normal}$ & $AP_{defect}$& \revise{Params} & \revise{FLOPs} & \revise{FPS} \\ \hline
        \multirow{2}{*}{ResNet 34} & Detection & $97.70$ & $98.23$ & $98.23$ & $92.89$ & $98.02$ & $97.38$ & \multirow{2}{*}{\revise{41M}} & \multirow{2}{*}{\revise{274G}} & \multirow{2}{*}{\revise{5.1}} \\ 
        ~ & Segmentation & $99.22$ & $99.22$ & $99.22$ & $99.22$ & $100.00$ & $98.45$ & ~ & ~ & ~ \\ \midrule
        \multirow{2}{*}{ResNet 50} & Detection & $95.18$ & $95.18$ & $95.18$ & $95.18$ & $94.02$ & $96.34$ & \multirow{2}{*}{\revise{41M}} & \multirow{2}{*}{\revise{274G}} & \multirow{2}{*}{\revise{5.1}} \\ 
        ~ & Segmentation & $98.43$ & $98.43$ & $98.43$ & $98.43$ & $99.79$ & $97.06$ & ~ & ~ & ~ \\ \midrule
        \  \multirow{2}{*}{\revise{Swin-T}} & \revise{Detection} & \revise{$95.69$} & \revise{$95.79$} & \revise{$95.93$} & \revise{$95.32$} & \revise{$94.23$} & \revise{$97.15$} & \multirow{2}{*}{\revise{46M}} & \multirow{2}{*}{\revise{281G}} & \multirow{2}{*}{\revise{4.5}} \\ 
        ~ & \revise{Segmentation} & \revise{$97.77$} & \revise{$98.65$} & \revise{$98.93$} & \revise{$98.15$} & \revise{$99.11$} & \revise{$96.42$} & ~ & ~ & ~ \\ \hline
    \end{tabular}
    }
    \label{tab:detail_result_single}
\end{table}


\begin{table}[!h]
    \centering
    \caption{Detailed Performance of CarcassFormer on Multiple Carcasses Per Image Dataset on both Detection and Segmentation, whereas $AP_{normal}$ \& $AP_{defect}$ include classification results. }
     \renewcommand{\arraystretch}{1.3}
    \setlength{\tabcolsep}{4pt}
    \resizebox{1.0\textwidth}{!}{
    \begin{tabular}{c|c|cccccc|ccc}
    \hline
        \textbf{Backbone} & \textbf{Task} & $AP$ & $AP_{50}$ & $AP_{75}$ & $AP_{95}$ & $AP_{normal}$ & $AP_{defect}$ & \revise{Params} & \revise{FLOPs} & \revise{FPS}\\ \hline
        \multirow{2}{*}{ResNet 34} & Detection & $89.72$ & $91.45$ & $91.45$ & $78.51$ & $93.48$ & $85.96$ & \multirow{2}{*}{\revise{41M}} & \multirow{2}{*}{\revise{274G}} & \multirow{2}{*}{\revise{5.1}} \\ 
        ~ & Segmentation & $98.23$ & $99.34$ & $98.86$ & $92.55$ & $98.77$ & $97.68$ & ~ & ~ & ~ \\ \midrule
        \multirow{2}{*}{ResNet 50} & Detection & $90.45$ & $91.55$ & $91.55$ & $83.41$ & $93.42$ & $87.49$ & \multirow{2}{*}{\revise{41M}} & \multirow{2}{*}{\revise{274G}} & \multirow{2}{*}{\revise{5.1}} \\ 
        ~ & Segmentation & $98.96$ & $99.98$ & $99.48$ & $94.36$ & $99.15$ & $98.76$ & ~ & ~ & ~ \\ \midrule
        \multirow{2}{*}{\revise{Swin-T}} & \revise{Detection} & \revise{$89.34$} & \revise{$91.27$} & \revise{$91.73$} & \revise{$79.11$} & \revise{$94.18$} & \revise{$84.50$}  & \multirow{2}{*}{\revise{46M}} & \multirow{2}{*}{\revise{281G}} & \multirow{2}{*}{\revise{4.5}} \\ 
        ~ & \revise{Segmentation} & \revise{$98.70$} & \revise{$99.29$} & \revise{$98.32$} & \revise{$93.10$} & \revise{$99.47$} & \revise{$97.92$} & ~ & ~ & ~ \\ \hline
    \end{tabular}
    }
    \label{tab:detail_result_overlap}
\end{table}

\subsubsection{Detailed Quantitative Performance}

Detailed performance conducted by our CarcassFormer is reported in Table \ref{tab:detail_result_single} and Table \ref{tab:detail_result_overlap} corresponding to two subsets: a single carcass per image and multiple carcasses per image. In each subset, our CarcassFormer network was examined on \revise{three} different backbone networks: ResNet-34, ResNet-50, \revise{and Swin-T}. For both tasks of detection and segmentation, Average Precision (AP) at different metrics of AP@50, AP@75, AP@95, and AP[50:95] (referred to as AP) were reported. Regarding detection and classification, $AP_{normal}$ and $AP_{defect}$ were evaluated for normal and defect classes. 
\revise{The results obtained from two tables (Table \ref{tab:detail_result_single} and \ref{tab:detail_result_overlap}) underscores the remarkable performance of our model across various backbones and tasks, with every configuration achieving an AP of over 85 for all metrics. Additionally, it becomes evident that the Multiple Carcasses Per Image Dataset presents greater challenges compared to the Single Carcasses Per Image Dataset. This observation is substantiated by a noticeable decline in performance metrics when handling multiple overlapping carcasses per image, as opposed to the single carcass per image scenario.}

\begin{table}[!h]
    \centering
    \caption{Performance comparison between CarcassFormer with both CNN-based networks, namely Mask R-CNN \cite{he2017mask} and HTC \cite{chen2019hybrid} and Transformer-based networks, namely Mask2Former \cite{cheng2022masked} and QueryInst \cite{fang2021instances} on both Detection, Classification and Segmentation tasks. The comparison is conducted on \textbf{ResNet-34} backbone network and in the case of \textbf{single carcass per image}. Net. denotes Network architecture}
    \renewcommand{\arraystretch}{1.3}
    \setlength{\tabcolsep}{3pt}
    \resizebox{\textwidth}{!}{
    \begin{tabular} {l|l|c|ccc|ccc|ccc}
    \toprule
        \multirow{2}{*}{\textbf{Net.}} & \multirow{2}{*}{\textbf{Method}} & \multirow{2}{*}{\textbf{Venue}} & \multicolumn{3}{c}{\shortstack{\textbf{Detection} \\ \& \textbf{Classification}}} & \multicolumn{3}{|c}{\textbf{Segmentation}} &  \multicolumn{3}{|c}{\revise{\textbf{\shortstack{Model \\ Complexity}}}}\\ \cline{4-12}
        ~ & ~ & ~ & $AP$ & $AP_{normal}$ & $AP_{defect}$ & $AP$ & $AP_{normal}$ & $AP_{defect}$ & \revise{Params} &\revise{FLOPs} & \revise{FPS} \\ \midrule
\multirow{2}{*}{\rotatebox{90}{\shortstack{CNN\\-based}}} & Mask R-CNN \cite{he2017mask} & $2017$ & $79.73$ & $87.82$ & $71.65$ & $81.36$ & $85.41$ & $77.31$ & \revise{41M} & \revise{204G} & \revise{5.8} \\ 
        ~ & HTC \cite{chen2019hybrid} & $2019$ & $89.00$ & $95.30$ & $82.60$ & $82.30$ & $86.30$ & $78.40$ & \revise{109M} & \revise{290G} & \revise{4.3} \\  \cmidrule{2-12}
        \multirow{4}{*}{\rotatebox{90}{\shortstack{Transformer\\-based}}} & QueryInst \cite{fang2021instances} & $2021$ & $90.40$ & $98.00$ & $82.90$ & $82.20$ & $88.10$ & $76.30$ & \revise{45M} & \revise{279G} & \revise{4.8} \\ 
        ~ & Mask2Former \cite{cheng2022masked} & $2022$ & $58.33$ & $58.42$ & $58.24$ & $75.32$ & $92.08$ & $58.57$ & \revise{41M} & \revise{272G} & \revise{5.2} \\ 
        ~ & \revise{Mask DINO \cite{li2023mask} } & \revise{2023} & \revise{88.12} & \revise{95.11} & \revise{81.12} & \revise{77.67} & \revise{83.45} & \revise{71.89} & \revise{49M} & \revise{278G} & \revise{5.1} \\ \cmidrule{2-12}
        ~ & \textbf{CarcassFormer} (Ours) & ~ & $\mathbf{97.70}$ & $\mathbf{98.02}$ & $\mathbf{97.38}$ & $\mathbf{99.22}$ & $\mathbf{100.00}$ & $\mathbf{98.45}$ & \revise{41M} & \revise{274G} & \revise{5.1} \\ \hline
    \end{tabular}
    } 
\label{tab:single_r34}
\end{table}

\begin{table}
    \centering
    \caption{Performance comparison between CarcassFormer with both CNN-based networks, namely Mask R-CNN \cite{he2017mask} and HTC \cite{chen2019hybrid} and Transformer-based networks, namely Mask2Former \cite{cheng2022masked} and QueryInst \cite{fang2021instances} on both Detection, Classification and Segmentation tasks. The comparison is conducted on \textbf{ResNet-50} backbone network and in the case of \textbf{single carcass per image}. Net. denotes Network architecture. The best score in each table is highlighted in \textbf{bold}.}
    \renewcommand{\arraystretch}{1.3}
    \setlength{\tabcolsep}{3pt}
    \resizebox{\textwidth}{!}{
    \begin{tabular} {l|l|c|ccc|ccc|ccc}
    \toprule
        \multirow{2}{*}{\textbf{Net.}} & \multirow{2}{*}{\textbf{Method}} & \multirow{2}{*}{\textbf{Venue}} & \multicolumn{3}{c}{\shortstack{\textbf{Detection} \\ \& \textbf{Classification}}} & \multicolumn{3}{|c}{\textbf{Segmentation}} &  \multicolumn{3}{|c}{\revise{\textbf{\shortstack{Model \\ Complexity}}}}\\ \cline{4-12}
        ~ & ~ & ~ & $AP$ & $AP_{normal}$ & $AP_{defect}$ & $AP$ & $AP_{normal}$ & $AP_{defect}$ & \revise{Params} &\revise{FLOPs} & \revise{FPS} \\ \midrule
\multirow{2}{*}{\rotatebox{90}{\shortstack{CNN\\-based}}} & Mask R-CNN \cite{he2017mask} & $2017$ & $80.35$ & $90.15$ & $70.56$ & $84.19$ & $88.39$ & $80.00$ & \revise{44M} & \revise{207G} & \revise{5.3} \\ 
        ~ & HTC \cite{chen2019hybrid} & $2019$ & $88.10$ & $\mathbf{96.20}$ & $80.00$ & $84.30$ & $89.00$ & $79.70$ & \revise{112M} & \revise{294G} & \revise{3.9} \\ \cmidrule{2-12}
        \multirow{4}{*}{\rotatebox{90}{\shortstack{Transformer\\-based}}} & QueryInst \cite{fang2021instances} & $2021$ & $64.60$ & $75.50$ & $53.60$ & $72.70$ & $78.60$ & $66.70$ & \revise{48M} & \revise{281G} & \revise{4.4} \\ 
        ~ & Mask2Former \cite{cheng2022masked} & $2022$ & $85.05$ & $91.23$ & $78.87$ & $85.11$ & $91.23$ & $78.99$ & \revise{44M} & \revise{276G} & \revise{4.8} \\ 
        ~ & \revise{Mask DINO \cite{li2023mask} } & \revise{2023} & \revise{85.12} & \revise{91.44} & \revise{79.10} & \revise{86.13} & \revise{92.11} & \revise{80.15} & \revise{52M} & \revise{280G} & \revise{4.6} \\ \cmidrule{2-12}
        ~ & CarcassFormer (Ours) & ~ & $\mathbf{95.18}$ & $94.02$ & $\mathbf{96.34}$ & $\mathbf{98.43}$ & $\mathbf{99.79}$ & $\mathbf{97.06}$ & \revise{44M} & \revise{278G} & \revise{4.6} \\ \hline
    \end{tabular}
    } 
\label{tab:single_r50}
\end{table}

\subsubsection{Single Carcass Per Image}

Table \ref{tab:single_r34} and Table \ref{tab:single_r50} present the performance on a single carcass per image using ResNet-34 and ResNet-50, respectively.

Table \ref{tab:single_r34} compares the performance of CarcassFormer with existing methods on ResNet-34. In the first group, HTC \cite{chen2019hybrid} obtains better performance than Mask R-CNN \cite{he2017mask} whereas our CarcassFormer gains significant performance gaps compared to both HTC \cite{chen2019hybrid} and Mask R-CNN \cite{he2017mask}. Take HTC \cite{chen2019hybrid} as an example, CarcassFormer outperforms HTC with 8.70\% higher AP for detection, 2.72\% higher AP for normal carcass classification, 14.78 \% higher AP for defect carcass classification, 16.92\% higher AP for segmentation, 13.70\% higher AP segmentation for normal carcass and 20.05\% higher AP segmentation for defect carcass. In the second group, QueryInst \cite{fang2021instances} obtains better performance than Mask2Former  \cite{cheng2022masked} \revise{and Mask DINO~\cite{li2023mask}} while our CarcassFormer obtains the best performance. Compared to QueryInst \cite{fang2021instances}, CarcassFormer gains 7.30\% higher AP for detection, 0.02\% higher AP for normal carcass classification 14.48\% higher AP for defect carcass classification, 17.02\% higher AP for segmentation, 11.90\% higher AP segmentation for normal carcass, 22.15\% higher AP segmentation for defect carcass.

Table \ref{tab:single_r50} compares the performance of CarcassFormer with existing methods on ResNet-50. In the first group, HTC \cite{chen2019hybrid} outperforms Mask R-CNN \cite{he2017mask} whereas CarcassFormer outperforms HTC \cite{chen2019hybrid} with significant performance gaps, including 7.08\% higher AP for detection, 16.34\% higher AP for defect carcass classification, 14.13\% higher AP for segmentation, 10.79\% higher AP segmentation for normal carcass and 17.36\% higher AP segmentation for defect carcass while it is compatible with HTC \cite{chen2019hybrid} on normal classification. In the second group, while Mask2Former \cite{cheng2022masked} \revise{and Mask DINO~\cite{li2023mask}} obtains much better performance than QueryInst \cite{fang2021instances}, \revise{our CarcassFormer outperforms MaskDINO~\cite{li2023mask} 10.06\% higher AP for detection, 2.58\% higher AP for normal carcass classification, 17.24\% higher AP for defect carcass classification,  13.30\% higher AP for segmentation, 7.68\% higher AP segmentation for normal carcass, 16.91\% higher AP segmentation for defect carcass.}


\begin{table}[!h]
    \centering
    \caption{Performance comparison between CarcassFormer with both CNN-based networks, namely Mask R-CNN \cite{he2017mask} and HTC \cite{chen2019hybrid} and Transformer-based networks, namely Mask2Former \cite{cheng2022masked} and QueryInst \cite{fang2021instances} on both Detection, Classification and Segmentation tasks. The comparison is conducted on \textbf{ResNet-34} backbone network and in the case of \textbf{multiple carcasses per image}. Net. denotes Network architecture. The best score in each table is highlighted in \textbf{bold}.}
    \renewcommand{\arraystretch}{1.3}
    \setlength{\tabcolsep}{3pt}
    \resizebox{\textwidth}{!}{
    \begin{tabular} {l|l|c|ccc|ccc|ccc}
    \toprule
        \multirow{2}{*}{\textbf{Net.}} & \multirow{2}{*}{\textbf{Method}} & \multirow{2}{*}{\textbf{Venue}} & \multicolumn{3}{c}{\shortstack{\textbf{Detection} \\ \& \textbf{Classification}}} & \multicolumn{3}{|c}{\textbf{Segmentation}} &  \multicolumn{3}{|c}{\revise{\textbf{\shortstack{Model \\ Complexity}}}}\\ \cline{4-12}
        ~ & ~ & ~ & $AP$ & $AP_{normal}$ & $AP_{defect}$ & $AP$ & $AP_{normal}$ & $AP_{defect}$ & \revise{Params} &\revise{FLOPs} & \revise{FPS} \\ \midrule
        \multirow{2}{*}{\rotatebox{90}{\shortstack{CNN\\-based}}} & Mask R-CNN \cite{he2017mask} & $2017$ & $77.08$ & $84.33$ & $69.83$ & $74.81$ & $79.00$ & $70.63$ & \revise{41M} & \revise{204G} & \revise{5.8} \\ 
        ~ & HTC \cite{chen2019hybrid} & $2019$ & $77.80$ & $89.70$ & $65.90$ & $74.00$ & $79.10$ & $68.90$ & \revise{109M} & \revise{290G} & \revise{4.3} \\ \cmidrule{2-12}
        \multirow{4}{*}{\rotatebox{90}{\shortstack{Transformer\\-based}}} & QueryInst \cite{fang2021instances} & $2021$ & $84.10$ & $89.60$ & $78.70$ & $83.20$ & $87.70$ & $78.70$ & \revise{45M} & \revise{279G} & \revise{4.8} \\ 
        ~ & Mask2Former \cite{cheng2022masked} & $2022$ & $53.86$ & $54.00$ & $53.72$ & $71.69$ & $85.39$ & $58.00$ & \revise{41M} & \revise{272G} & \revise{5.2} \\ 
         ~ & \revise{Mask DINO \cite{li2023mask} } & \revise{2023} & \revise{68.44} & \revise{74.55} & \revise{62.33} & \revise{78.80} & \revise{88.26} & \revise{69.33} & \revise{49M} & \revise{278G} & \revise{5.1} \\ \cmidrule{2-12}
        ~ & \textbf{CarcassFormer} (Ours) & ~ & $\textbf{89.72}$ & $\textbf{93.48}$ & $\textbf{85.96}$ & $\textbf{98.23}$ & $\textbf{98.77}$ & $\textbf{97.68}$ & \revise{41M} & \revise{274G} & \revise{5.1} \\ \hline
    \end{tabular}
    } 
    \label{tab:compare_overlap_r34}
    
\end{table}

\begin{table}[!h]
    \centering
    \caption{Performance comparison between CarcassFormer with both CNN-based networks, namely Mask R-CNN \cite{he2017mask} and HTC \cite{chen2019hybrid} and Transformer-based networks, namely Mask2Former \cite{cheng2022masked} and QueryInst \cite{fang2021instances} on both Detection, Classification and Segmentation tasks. The comparison is conducted on \textbf{ResNet-50} backbone network and in the case of \textbf{multiple carcasses per image}. Net. denotes Network architecture. The best score in each table is highlighted in \textbf{bold}.}
    \renewcommand{\arraystretch}{1.3}
    \setlength{\tabcolsep}{3pt}
    \resizebox{\textwidth}{!}{
    \begin{tabular} {l|l|c|ccc|ccc|ccc}
    \toprule
        \multirow{2}{*}{\textbf{Net.}} & \multirow{2}{*}{\textbf{Method}} & \multirow{2}{*}{\textbf{Venue}} & \multicolumn{3}{c}{\shortstack{\textbf{Detection} \\ \& \textbf{Classification}}} & \multicolumn{3}{|c}{\textbf{Segmentation}} &  \multicolumn{3}{|c}{\revise{\textbf{\shortstack{Model \\ Complexity}}}}\\ \cline{4-12}
        ~ & ~ & ~ & $AP$ & $AP_{normal}$ & $AP_{defect}$ & $AP$ & $AP_{normal}$ & $AP_{defect}$ & \revise{Params} &\revise{FLOPs} & \revise{FPS} \\ \midrule
        \multirow{2}{*}{\rotatebox{90}{\shortstack{CNN\\-based}}} & Mask R-CNN \cite{he2017mask} & $2017$ & $78.76$ & $85.61$ & $75.73$ & $80.67$ & $85.61$ & $75.73$ & \revise{44M} & \revise{207G} & \revise{5.3} \\ 
        ~ & HTC \cite{chen2019hybrid} & $2019$ & $77.40$ & $83.50$ & $71.40$ & $74.90$ & $77.70$ & $72.10$ & \revise{112M} & \revise{294G} & \revise{3.9} \\ \cmidrule{2-12}
        \multirow{4}{*}{\rotatebox{90}{\shortstack{Transformer\\-based}}} & QueryInst \cite{fang2021instances} & $2021$ & $60.90$ & $67.70$ & $54.00$ & $60.40$ & $66.90$ & $54.00$ & \revise{48M} & \revise{281G} & \revise{4.4} \\ 
        ~ & Mask2Former \cite{cheng2022masked} & $2022$ & $73.35$ & $88.03$ & $58.66$ & $75.54$ & $90.93$ & $60.15$ & \revise{44M} & \revise{276G} & \revise{4.8} \\ 
        ~ & \revise{Mask DINO \cite{li2023mask} } & \revise{2023} & \revise{76.22} & \revise{84.11} & \revise{68.33} & \revise{79.93} & \revise{92.12} & \revise{67.74} & \revise{52M} & \revise{280G} & \revise{4.6} \\ \cmidrule{2-12}
        ~ & \textbf{CarcassFormer} (Ours) & ~ & $\textbf{90.45}$ & $\textbf{93.42}$ & $\textbf{87.49}$ & $\textbf{98.96}$ & $\textbf{99.15}$ & $\textbf{98.76}$ & \revise{44M} & \revise{278G} & \revise{4.6} \\ \hline
    \end{tabular}
    } 
    \label{tab:compare_overlap_r50}

\end{table}

\begin{table}[!h]
    \centering
    \caption{\revise{Performance comparison between CarcassFormer with Mask2Former \cite{cheng2022masked} on both Detection, Classification and Segmentation tasks. The comparison is conducted on \textbf{Swin-T} backbone network and in the case of \textbf{multiple carcasses per image}. Net. denotes Network architecture. The best score in each table is highlighted in \textbf{bold}.}}
    \renewcommand{\arraystretch}{1.3}
    \setlength{\tabcolsep}{3pt}
    \resizebox{\textwidth}{!}{
    \begin{tabular} {l|l|c|ccc|ccc|ccc}
    \toprule
         & \multirow{2}{*}{\revise{\textbf{Method}}} & \multirow{2}{*}{\revise{\textbf{Venue}}} & \multicolumn{3}{c}{\shortstack{\revise{\textbf{Detection}} \\ \revise{\&} \revise{\textbf{Classification}}}} & \multicolumn{3}{|c}{\revise{\textbf{Segmentation}}} &  \multicolumn{3}{|c}{\revise{\textbf{\shortstack{Model \\ Complexity}}}}\\ \cline{4-12}
         ~ & ~ & ~ & \revise{$AP$} & \revise{$AP_{normal}$} & \revise{$AP_{defect}$} & \revise{$AP$} & \revise{$AP_{normal}$} & \revise{$AP_{defect}$} & \revise{Params} &\revise{FLOPs} & \revise{FPS} \\ \midrule
        ~ & \revise{Mask R-CNN \cite{he2016deep}} & \revise{$2017$} & \revise{$78.82$} & \revise{$86.12$} & \revise{$71.52$} & \revise{$81.22$} & \revise{$87.12$} & \revise{$75.32$} & \revise{46M} & \revise{230G} & \revise{4.8} \\ 
        ~ & \revise{Mask2Former \cite{cheng2022masked}} & \revise{$2022$} & \revise{$73.10$} & \revise{$88.14$} & \revise{$58.05$} & \revise{$77.68$} & \revise{$92.89$} & \revise{$62.47$} & \revise{46M} & \revise{280G} & \revise{4.6} \\ 
        ~ & \revise{\textbf{CarcassFormer} (Ours)} & ~ & \revise{$\textbf{89.34}$} & \revise{$\textbf{94.18}$} & \revise{$\textbf{84.50}$} & \revise{$\textbf{98.70}$} & \revise{$\textbf{99.47}$} & \revise{$\textbf{97.92}$} & \revise{46M} & \revise{281G} & \revise{4.5} \\ \hline
    \end{tabular}
    }
    \label{tab:compare_overlap_swint}
\end{table}

\subsubsection{Multiple Carcasses Per Image}

Table \ref{tab:compare_overlap_r34} and Table \ref{tab:compare_overlap_r50} present the performance on multiple carcasses per image using ResNet-34 and ResNet-50, respectively.

Table \ref{tab:compare_overlap_r34} compares the performance of CarcassFormer with existing methods on ResNet-34. In the first group, while Mask R-CNN \cite{he2017mask} and HTC \cite{chen2019hybrid} are quite compatible on all tasks, our CarcassFormer gains big performance gaps. Take HTC \cite{chen2019hybrid} as an example, CarcassFormer achieves 11.92\% higher AP for detection, 3.78\% higher AP for normal carcass classification, 20.06\% higher AP for defect carcass classification, 24.23\% higher AP for segmentation, 19.67\% higher AP segmentation for normal carcass, 28.79\% higher AP segmentation for defect carcass. In the second group, while QueryInst \cite{fang2021instances} outperforms Mask2Former \cite{cheng2022masked} \revise{and Mask DINO~\cite{li2023mask}}, our CarcassFormer obtains better performance than QueryInst \cite{fang2021instances} with notable gaps, i.e., 5.62\% higher AP for detection, 3.88\% higher AP for normal carcass classification, 7.26\% higher AP for defect carcass classification, 15.03\% higher AP for segmentation, 11.07\% higher AP segmentation for normal carcass, 18.98\% higher AP segmentation for defect carcass.

Table \ref{tab:compare_overlap_r50} compares the performance of CarcassFormer with existing methods on ResNet-50. In the first group, while Mask R-CNN \cite{he2017mask} outperforms HTC \cite{chen2019hybrid}, our CarcassFormer achieves a best performance with 11.69\% higher AP for detection, 7.81\% higher AP for normal carcass classification, 11.76\% higher AP for defect carcass classification, 18.29\% higher AP for segmentation, 13.54\% higher AP segmentation for normal carcass, 23.03\% higher AP segmentation for defect carcass compared to Mask R-CNN \cite{he2017mask}. In the second group, while Mask2Former \cite{cheng2022masked} \revise{Mask DINO~\cite{li2023mask}} obtain better performance than QueryInst \cite{fang2021instances}, our CarcassFormer achieves the best performance. It gains \revise{14.23\%} higher AP for detection, \revise{9.31\%} higher AP for normal carcass classification, \revise{19.16\%} higher AP for defect carcass classification, \revise{19.03\%} higher AP for segmentation, \revise{7.03\%} higher AP segmentation for normal carcass, \revise{31.02\%} higher AP segmentation for defect carcass compared to the second best method \revise{Mask DINO \cite{li2023mask}.}

\revise{Table \ref{tab:compare_overlap_swint} compares the performance of CarcassFormer with Mask2Former~\cite{cheng2022masked} and Mask R-CNN~\cite{he2017mask} on Swin-T. In comparison to the CNN-based method, Mask R-CNN~\cite{he2017mask}, our approach yields significant improvements across various performance metrics. Specifically, we observe a 10.52\% increase in average precision (AP) for detection, an 8.06\% enhancement for normal carcass classification, a 12.98\% boost for defect carcass classification, a remarkable 17.48\% rise for segmentation, as well as notable gains of 12.35\% and 22.6\% in AP segmentation for normal and defect carcasses, respectively.
In terms of comparison with transformer based network, namely Mask2Former, our CarcassFormer achieves the significant better performance than both Mask2Former and Mask R-CNN. Indeed, it gains 16.24\% higher AP for detection, 6.04\% higher AP for normal carcass classification, 26.45\% higher AP for defect carcass classification, 21.02\% higher AP for segmentation, 6.58\% higher AP segmentation for normal carcass, 35.45\% higher AP segmentation for defect carcass. }\\

\noindent
\revise{\textbf{Model Complexity.} Analysis of model complexity reveals that our method exhibits comparable complexity to the majority of existing methods. However, it consistently delivers notable performance enhancements across diverse tasks. 
Specifically, in the case of ResNet-34, as illustrated in Tables \ref{tab:single_r34} and \ref{tab:compare_overlap_r34}, our model has the smallest number of model parameters, equivalent to that of Mask R-CNN~\cite{he2017mask} and Mask2Former~\cite{cheng2022masked}, while maintaining comparable FLOPs and FPS with these models. However, our model exhibits a significant performance advantage over both. This trend is similarly observed for ResNet-50, as shown in Tables \ref{tab:single_r50} and \ref{tab:compare_overlap_r50}, and for Swin-T, as depicted in Table~\ref{tab:compare_overlap_swint}, where our model demonstrates comparable model complexity but yields substantial performance gains compared to other methods.}

\begin{figure}[!h] 
    \centering
    \includegraphics[width=\linewidth]{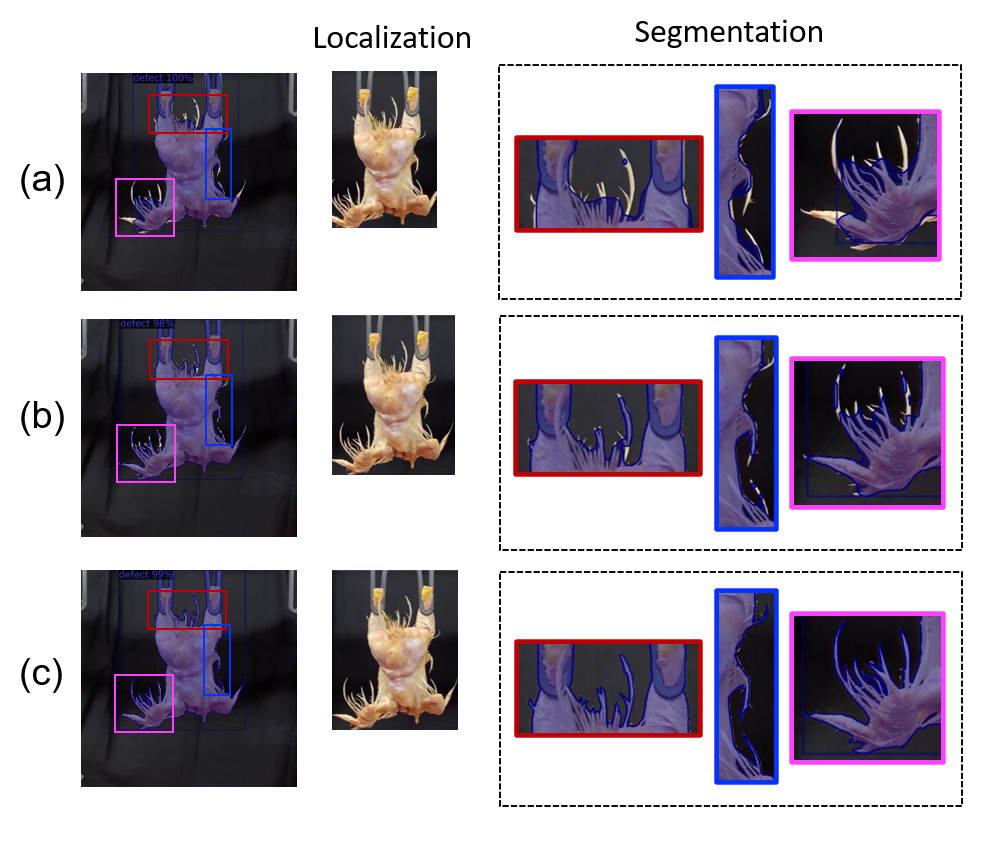}
    \caption{Performance comparison (a): Mask R-CNN \cite{he2016deep}, (b): Mask2Former \cite{cheng2022masked} and (c): our CarcassFormer on the defect where \textbf{single carcass with feathers}. In the Segmentation column, notable parts with feathers were highlighted. Compare with Mask R-CNN and Mask2Former, our CarcassFormer can localize carcass with more accurate bounding box and segment carcass with more details on feathers.}
    \label{fig:qual_defect1}
\end{figure}

\begin{figure}[!h] 
    \centering
    \includegraphics[width=0.9\linewidth]{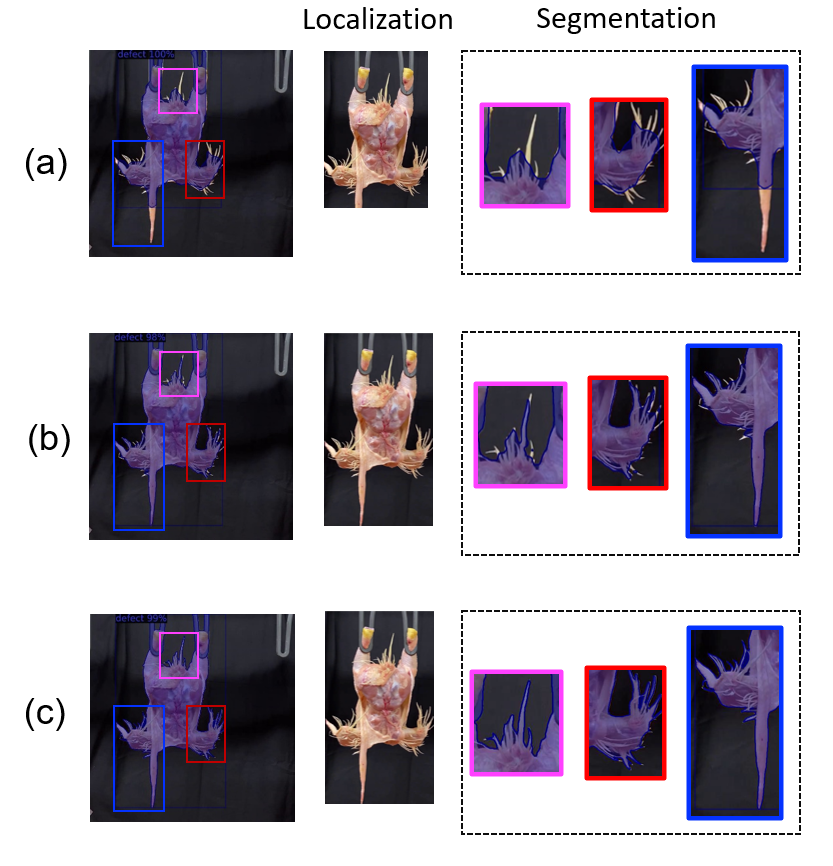}
    \caption{Performance comparison (a): Mask R-CNN \cite{he2016deep}, (b): Mask2Former \cite{cheng2022masked} and (c): our CarcassFormer on two defects \textbf{where single carcass with skins tearing on the back and feathers}. In the Segmentation column, notable parts with feathers and skins tearing occurred were highlighted. Compared with Mask R-CNN and Mask2Former, our CarcassFormer desnt not only detect the feathers well but also accurately localize carcass with its all skins tearing.}
    \label{fig:qual_defect2}
\end{figure}
\begin{figure}[!h] 
    \centering
    \includegraphics[width=0.9\linewidth]{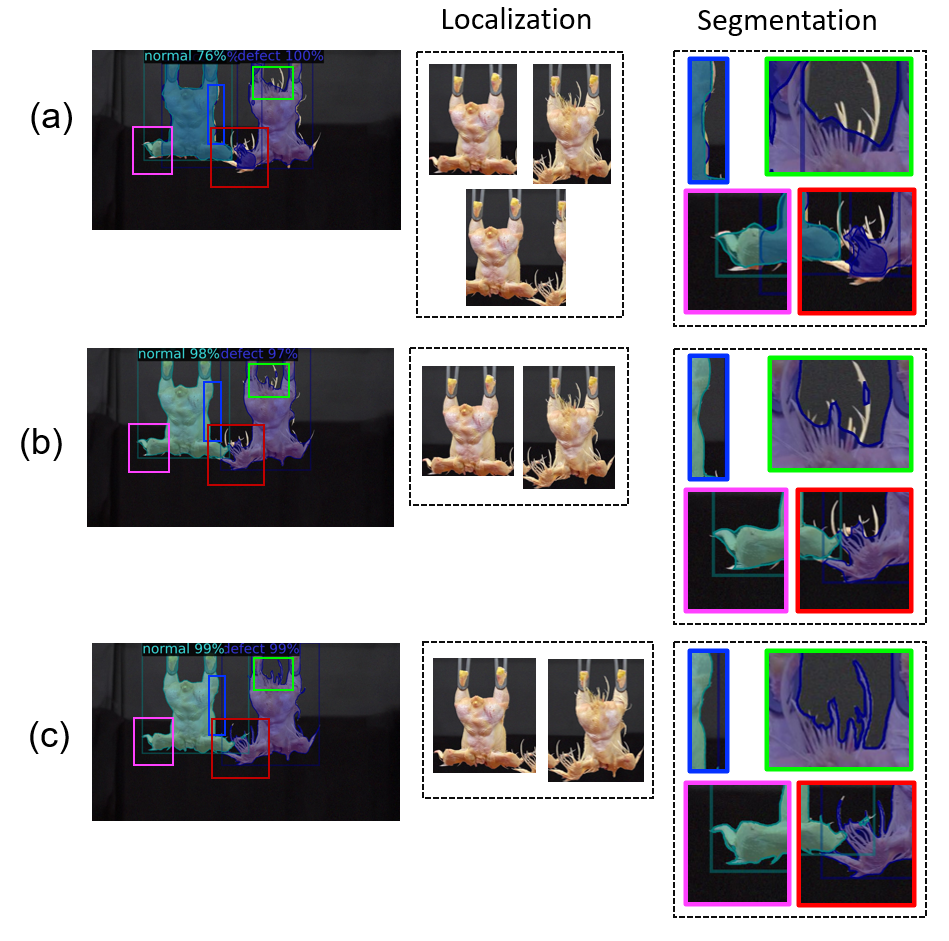}
    \caption{Performance comparison between Mask R-CNN (a), \cite{he2016deep} Mask2Former \cite{cheng2022masked} (b) and our CarcassFormer (c) on \textbf{overlapping carcasses with feathers}. In the Segmentation column, notable parts with feathers were highlighted. Mask R-CNN not only lacks details in the segmentation results but also fails to localize individual carcasses. Although Mask2Former performs better than Mask R-CNN in localizing individual carcasses, it 
    still struggles to accurately segment all details. In contrast, our CarcassFormer can simultaneously segment carcasses with details and accurately localize individual carcasses.}
    \label{fig:qual_defect4}
\end{figure}
\begin{figure}[!h] 
    \centering
    \includegraphics[width=0.9\linewidth]{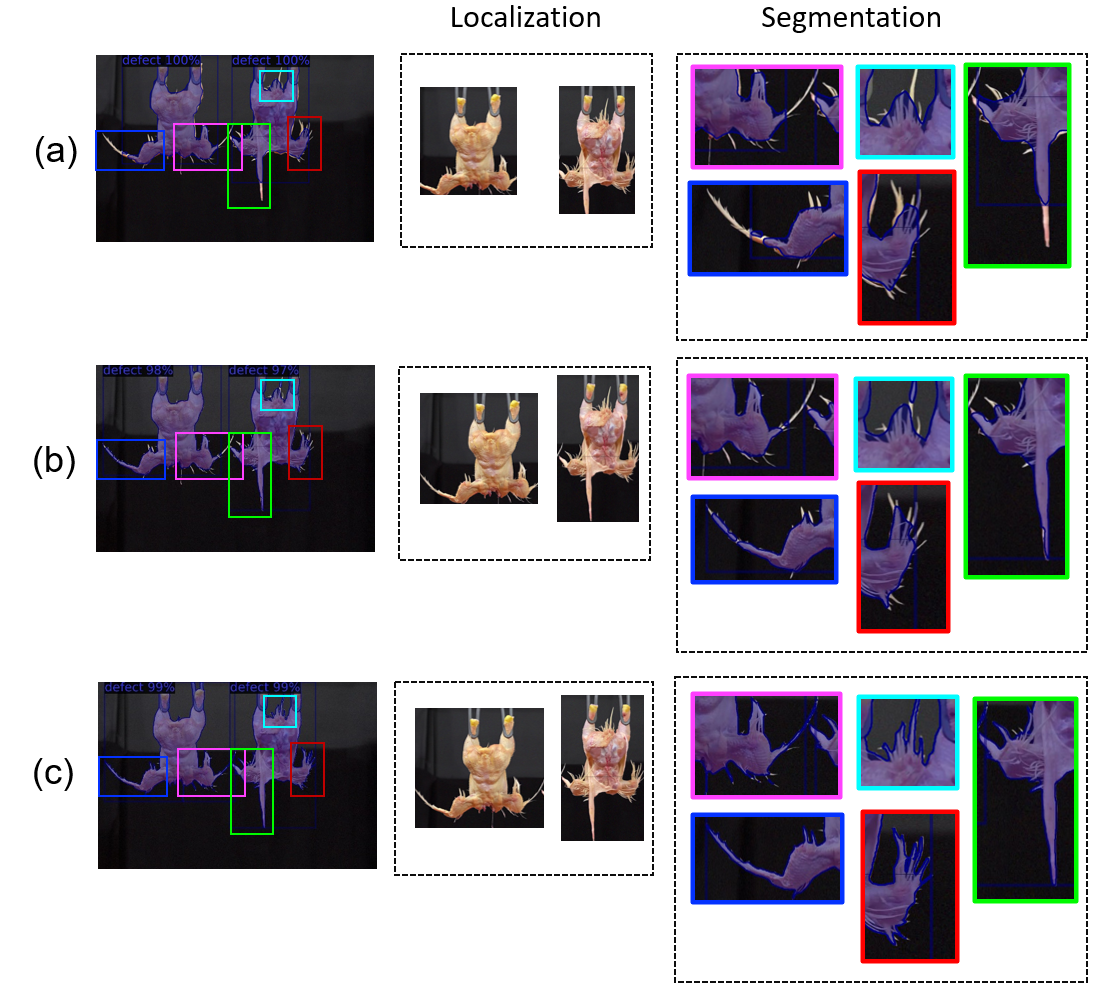}
    \caption{Performance comparison between (a): Mask R-CNN \cite{he2016deep}, (b): Mask2Former \cite{cheng2022masked} and (c): our CarcassFormer  on \textbf{overlapping carcasses with feathers and skins tearing}. Segmentation highlights some notable parts with feathers occur. Both Mask R-CNN and Mask2Former struggle with accurately localizing individual carcasses and providing detailed segmentation, especially for fine details like feathers. In contrast, our CarcassFormer excels in simultaneously segmenting carcasses with fine details and accurately localizing individual carcasses.}
    \label{fig:qual_defect3}
\end{figure}

\subsection{Qualitative Performance and Comparison}
\label{sec:performance_qual}

Based on the quantitative comparison in Section \ref{sec:performance}, Mask R-CNN \cite{he2017mask} was selected from the first group, and Mask2Former \cite{cheng2022masked} was chosen from the second group to conduct the qualitative comparison. Specifically, the qualitative comparison was reported on both the detection and segmentation tasks, with a greater emphasis on the case of defect, namely feather and skin tearing.

\subsubsection{Single Carcass Per Image}

Figure \ref{fig:qual_defect1} presents a qualitative performance comparison among three models: Mask R-CNN (a), Mask2Former (b), and our proposed CarcassFormer (c) on the defect of \textit{single carcass with feathers} is present. While Mask R-CNN can segment the global content well, it fails to segment the details, such as feathers. On the other hand, Mask2Former performs better than Mask R-CNN in capturing details, but it still faces difficulties in capturing fine details, which can be seen at high resolution. Moreover, Mask R-CNN and Mask2Former exhibited a tendency to under-localize the carcass, as observed from the detected bounding box that did not encompass the entire carcass with details on the boundary, such as wings and feathers. In contrast, our CarcassFormer not only accurately localizes the carcass with a fitting bounding box but is also capable of segmenting details at high resolution.

Figure \ref{fig:qual_defect2} depicts a qualitative performance comparison among three models: Mask R-CNN (a), Mask2Former (b), and our proposed CarcassFormer (c), on \textit{single carcass with both two defects of feather and skin tearing}. Although Mask2Former performs better than Mask R-CNN in localizing the carcass with skin tearing, it still faces difficulties in localizing all details on feathers. Conversely, our CarcassFormer accurately localizes the carcass with a fitting bounding box and is also capable of segmenting details at high resolution.

\subsubsection{Multiple Carcasses Per Image}

In this section, the qualitative performance on images with multiple carcasses and their overlap was reported. Figure \ref{fig:qual_defect4} illustrates the qualitative performance of three models: Mask R-CNN (a), Mask2Former (b), and our proposed CarcassFormer (c) \textit{on multiple, overlapping carcasses with feathers}. Both Mask R-CNN and Mask2Former struggle to accurately segment and localize each individual carcass, especially in cases where the feathers of different carcasses overlap. Mask R-CNN lacks detail in the segmentation results, while Mask2Former, despite performing better in localizing individual carcasses, still fails to capture all the details accurately. On the contrary, our CarcassFormer excels at accurately segmenting each individual carcass and capturing the details of feathers, even in complex scenarios of overlap.

In Figure \ref{fig:qual_defect3}, the qualitative performance comparison of the three models \textit{on multiple, overlapping carcasses with both feathers and skin tearing} presented. Here, it is evident that Mask R-CNN and Mask2Former both face significant challenges in accurately localizing individual carcasses and providing detailed segmentation, especially for tiny objects like feathers and areas of skin tearing. In stark contrast, our CarcassFormer performs outstandingly in these complex situations. It not only accurately localizes each carcass but also segments the fine details of feathers and skin tearing areas, thereby providing a comprehensive and detailed segmentation output.

\section{Conclusions}

In conclusion, an end-to-end Transformer-based network for checking carcass quality, CarcassFormer, has been described. Our CarcassFormer is designed with four different components: Network Backbone to extract visual features, Pixel Decoder to utilize feature maps from various scales, Mask-Attention Transformer Decoder to predict the segmented masks of all possible instances, and Instance Mask and Class Prediction to provide segmentation mask and corresponding label of an individual instance. To benchmark the proposed CarcassFormer network, a valuable realistic dataset was conducted at a poultry processing plant. The dataset acquired contained various defects including feathers, broken/disjointed bones, skins tearing, on different settings of a single carcass per image and multiple carcasses per image, and the carcass at various ages and sizes. The CarcassFormer was evaluated and compared with both CNN-based networks, namely Mask R-CNN \cite{he2017mask} and HTC \cite{chen2019hybrid}, as well as Transformer-based networks, namely Mask2Former \cite{cheng2022masked} and QueryInst \cite{fang2021instances}, on both detection, classification, and segmentation tasks using two different backbone networks, ResNet-34 and ResNet-50. The extensive qualitative and quantitative experiment showed that our CarcassFormer outperforms the existing methods with remarkable gaps on various metrics of AP, AP@50, AP@75. 

\revise{Our current CarcassFormer system operates solely on image-based inputs, limiting our ability to track carcasses across frames. While our model can currently determine whether a carcass is defective or not, it lacks the capability to identify specific types of defects, such as feathers around the carcass, feathers on the skin, flesh abnormalities, or broken wings.
In our future endeavors, we aim to expand our research to include video analysis, enabling us to track carcasses across frames and thereby enhance the scalability of our system to process a larger volume of carcasses. Additionally, we intend to implement finer-grained defect detection to precisely identify the nature of defects present. This enhancement will provide more detailed insights into the types of defects observed, facilitating improved diagnosis and statistical analysis.}

\section{Acknowledgments}

This material is based upon work supported by Cobb Vantress Inc., the National Science Foundation (NSF) under Award No OIA-1946391 RII Track-1, NSF 2119691 AI SUSTEIN.

\bibliographystyle{elsarticle-harv} 
\bibliography{cas-refs}





\end{document}